\documentclass[letterpaper, 10 pt, conference]{ieeeconf}  

\IEEEoverridecommandlockouts                              

\usepackage{verbatimbox}
\usepackage{color}
\usepackage{footnote}
\usepackage{bbding}
\usepackage{algorithmicx,algorithm}
\usepackage{algpseudocode}
\usepackage{graphicx}
\usepackage{subfigure}
\usepackage{amsfonts}
\usepackage{booktabs}
\usepackage{threeparttable}
\usepackage{babel}
\usepackage[table,xcdraw]{xcolor}
\usepackage{amsmath}
\usepackage{hyperref}
\hypersetup{
    colorlinks=true,
    linkcolor=blue,
    filecolor=blue,      
    urlcolor=blue,
    citecolor=cyan,
}

\usepackage{graphics} 
\usepackage{graphicx}
\usepackage{float}
\usepackage{pifont}       
\usepackage{bbding}       
\usepackage{fontawesome}  
\usepackage{multirow}
\usepackage{colortbl}  
\usepackage{graphicx}  
\usepackage{caption}   
\usepackage{arydshln}  
\usepackage{booktabs}  
\newcommand{\cmark}{\ding{51}}%
\newcommand{\xmark}{\text{\ding{55}}}
\newcommand{\grayrowcolor}{\rowcolor[gray]{0.85}}
\newcommand{\blurowcolor}{\rowcolor{blue!10}}
\definecolor{Blue}{RGB}{200,231,252}
\definecolor{Gray}{RGB}{210,210,210}

\title{\LARGE \bf
Pseudo Depth Meets Gaussian: A Feed-forward RGB SLAM Baseline
}

\author{
Linqing Zhao*, Xiuwei Xu*, Yirui Wang, Hao Wang, Wenzhao Zheng, Yansong Tang, Haibin Yan$^\dagger$, Jiwen Lu
\thanks{*Equal contribution. $^\dagger$Corresponding author.}
\thanks{
Linqing Zhao, Xiuwei Xu, Yirui Wang, Wenzhao Zheng, and Jiwen Lu are with the Department of Automation, Tsinghua University, Beijing 100084, China. Email: {zhaolinqing@mail.tsinghua.edu.cn}; {\{xxw21, wang-yr22\}@mails.tsinghua.edu.cn}; {wenzhao.zheng@outlook.com}; {lujiwen@tsinghua.edu.cn}.}
\thanks{
Hao Wang and Haibin Yan are with the School of Intelligent Engineering and Automation, Beijing University of Posts and Telecommunications, Beijing 100876, China. Email: \{2673439694, eyanhaibin\}@bupt.edu.cn.}
\thanks{
Yansong Tang is with the Shenzhen International Graduate School, Tsinghua University, Shenzhen 518071, China. Email: tang.yansong@sz.tsinghua.edu.cn.}
}

\begin{document}

\maketitle
\thispagestyle{empty}
\pagestyle{empty}

\begin{abstract}
Incrementally recovering real-sized 3D geometry from a pose-free RGB stream is a challenging task in 3D reconstruction, requiring minimal assumptions on input data. Existing methods can be broadly categorized into end-to-end and visual SLAM-based approaches, both of which either struggle with long sequences or depend on slow test-time optimization and depth sensors. To address this, we first integrate a depth estimator into an RGB-D SLAM system, but this approach is hindered by inaccurate geometric details in predicted depth. Through further investigation, we find that 3D Gaussian mapping can effectively solve this problem. Building on this, we propose an online 3D reconstruction method using 3D Gaussian-based SLAM, combined with a feed-forward recurrent prediction module to directly infer camera pose from optical flow. This approach replaces slow test-time optimization with fast network inference, significantly improving tracking speed. Additionally, we introduce a local graph rendering technique to enhance robustness in feed-forward pose prediction. Experimental results on the Replica and TUM-RGBD datasets, along with a real-world deployment demonstration, show that our method achieves performance on par with the state-of-the-art SplaTAM, while reducing tracking time by more than 90\%. Code is available at  \url{https://github.com/wangyr22/DepthGS}

\end{abstract}

\section{Introduction}
3D reconstruction aims to recover the 3D geometry from visual observations (e.g. RGB, RGB-D and stereo images) without camera poses, which is a fundamental and essential problem in computer vision. 
To facilitate real-world applications like robotics and AR / VR, we focus on a more challenging online 3D reconstruction setting, which incrementally reconstructs 3D point clouds of a scene given streaming video as input. The online 3D reconstruction requires on-the-fly construction of scene representation and tracking of camera pose, for which dense visual SLAM is a preferable solution.
However, classic methods rely on handcrafted representations such as points, surfels/flats, and signed distance fields. Constructing and tracking such representation highly depends on rich 3D feature extraction and high overlap between frames, which restricts the robustness of these approaches.

\begin{figure}[t]
    \centering
    \includegraphics[width=\linewidth]{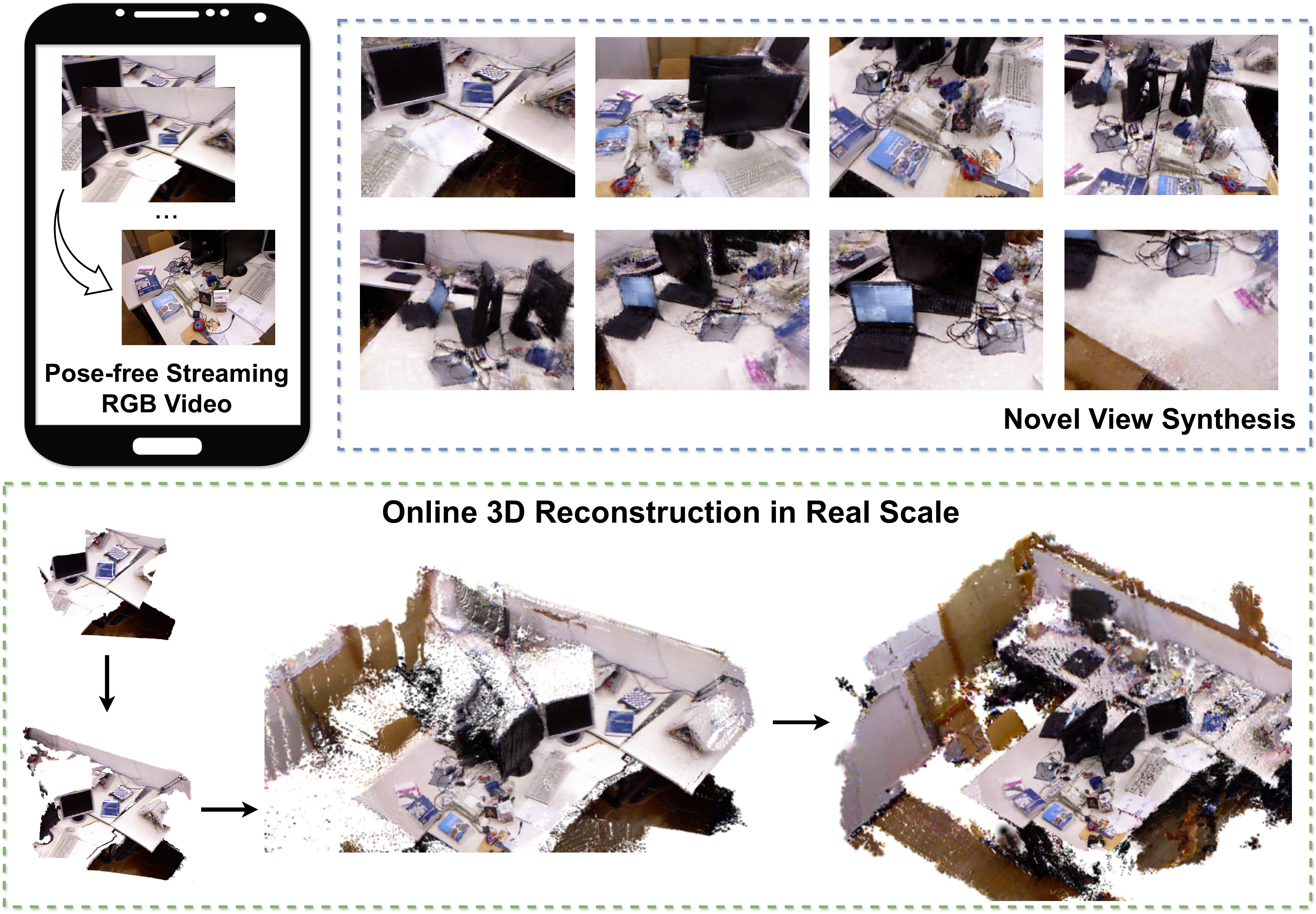} \\
    \caption{Given streaming RGB video without camera poses, our method can incrementally reconstruct frame-aligned 3D point clouds of the scene in real scale. Moreover, benefit from the 3D gaussian mapping, our method is also able to explain unobserved camera viewpoints with novel view synthesis.}
    \vspace{-5mm}
    \label{fig:teaser}
\end{figure}

With the emergence of neural radiance field~\cite{mildenhall2020nerf}, a recent class of methods leverage neural representation for scene mapping and camera tracking~\cite{kong2023vmap,sucar2021imap,wang2023co,zhu2022nice,sandstrom2023point}. Although these methods achieve robust 3D reconstruction by constructing high-fidelity global maps and capturing dense photometric information, they depend on volumetric ray sampling for optimization, which is very inefficient during rendering.
Adopt explicit 3D gaussian~\cite{kerbl20233d} as the SLAM representation~\cite{huang2024photo,keetha2024splatam,matsuki2024gaussian} can significantly speed up the rendering process, but they still require slow test-time optimization to refine 3D gaussian as well as solve camera poses. Moreover, these rendering-based dense visual SLAM methods rely on depth sensors like RGB-D cameras to acquire accurate 3D geometry, which is not feasible for a wide range of devices.
We also focus on this challenging setting, i.e., online, depth sensor-free, and fast 3D reconstruction. To achieve this, we build an efficient and robust framework upon a dense visual SLAM system to better handle large and complicated scenes.

We aim to build our system upon state-of-the-art RGB-D rendering-based SLAM method~\cite{keetha2024splatam}, while removing the requirement for depth sensor and replacing the time-consuming test-time optimization with feed-forward prediction.
In this way, our system can online reconstruct 3D point clouds given an RGB stream, while also being able to synthesize novel views thanks to the neural representation.
We start with a simple idea: leveraging depth estimator to convert RGB image to pseudo RGB-D image and then applying RGB-D SLAM for online 3D reconstruction. We empirically find this combination is actually infeasible for most RGB-D SLAM methods due to the inaccurate geometric details in the predicted depth. With in-depth study, we observe that 3D gaussian mapping can surprisingly solve this problem, which fully exploits the depth prior to construct scene representation and continuously refines it on-the-fly.
However, the camera pose tracking upon 3D gaussian mapping is extremely slow due to multiple iterations of pose optimization. To solve this problem, we propose a feed-forward pose prediction module for 3D gaussian, which renders a local graph for recurrent pose refinement prediction.
Specifically, given the 3D Gaussian representation of the scene and a target image (i.e. current RGB observation), we predict the approximate pose of the target image based on the inertia of the camera. We sample a series of viewpoints around the approximate pose and render images from 3D gaussian accordingly to construct a set of query images with known camera pose. Then we construct a local graph between query images and the target image and leverage a flow-based recurrent network to predict and refine the pose of the target image. This process is purely done by network inference, thus achieving much faster speed than optimization-based camera tracking methods.
Experimental results on the real-world TUM-RGBD and synthetic Replica datasets demonstrate the effectiveness of our approach. Our method achieves performance comparable to the state-of-the-art SLAM technique, SplaTAM, while reducing the tracking time by more than 90\%.

\section{Related Work}
\noindent \textbf{Classic 3D Reconstruction.}  
Recovering 3D geometry from 2D observations has been widely studied and can be divided into offline and online methods. Offline methods, like Structure-from-Motion~\cite{schonberger2016structure,sweeney2015optimizing}, reconstruct 3D scenes from a pre-collected image set by extracting sparse point clouds and camera poses through feature matching, geometric verification, and multi-view stereo~\cite{galliani2015massively,schonberger2016pixelwise,sun2021loftr}. However, offline methods are not suitable for real-time applications such as navigation~\cite{chaplot2020object}. Online methods, like visual SLAM, reconstruct the 3D scene on-the-fly using streaming video. Visual SLAM can be sparse~\cite{mur2015orb,engel2017direct} or dense~\cite{tateno2017cnn,zhou2018deeptam}, depending on the point cloud output. While existing online methods often rely on depth sensors like RGB-D cameras or LIDAR, they are sensitive to initialization and sub-task accuracy. In contrast, our method eliminates the need for depth sensors by using a depth estimator and leverages 3D Gaussian mapping for a robust global representation.

\noindent \textbf{End-to-end 3D Reconstruction.}  
SfM or SLAM-based 3D reconstruction pipelines involve multiple sub-tasks, making them vulnerable to noise. To address this, DUSt3R~\cite{wang2024dust3r} introduces an end-to-end 3D reconstruction approach that generates aligned 3D point clouds from two pose-free images. Follow-up works focus on feature matching~\cite{leroy2024grounding} and novel view synthesis~\cite{smart2024splatt3r}. However, DUSt3R is an offline method unsuitable for real-time tasks. Spann3R~\cite{wang20243d} extends this by incorporating spatial memory to enable online 3D reconstruction from pose-free RGB streams. Despite its potential, these methods struggle with long sequences due to memory issues. In contrast, our method uses explicit 3D Gaussian mapping with loop closure, ensuring robust performance over time.

\noindent \textbf{Neural Rendering-based 3D Reconstruction.}  
NeRF~\cite{mildenhall2020nerf} introduces a new 3D reconstruction paradigm using neural networks to represent a radiance field for novel view synthesis at any given pose. Follow-up works have applied NeRF's rendering technique for robust SfM~\cite{bian2023nope} and SLAM~\cite{kong2023vmap,sucar2021imap,wang2023co,zhu2022nice,hu2025cg,zhu2024nicer,peng2024rtg,li2024dngaussian}. Since NeRF is an implicit representation that is difficult to accumulate over time, Point-SLAM~\cite{sandstrom2023point} uses neural point cloud representation with volumetric rendering for better 3D reconstruction. However, these methods rely on time-consuming volumetric ray sampling for optimization. Recently, 3D Gaussian splatting~\cite{kerbl20233d} introduced an explicit representation and rendering framework, achieving superior accuracy and rendering speed. Variants of this approach in SLAM~\cite{huang2024photo,keetha2024splatam,matsuki2024gaussian,smith2024flowmap,zhang2023go} use 3D Gaussian as a mapping representation for real-time rendering. However, they still require test-time optimization, which limits inference speed, especially for tracking. To address this, we propose a feed-forward tracking method using local graph rendering and recurrent networks for camera pose prediction.

\begin{figure*}[t]
\centering
\subfigure[Mapping Representation Comparison]{
    \begin{minipage}{0.48\textwidth}
        \centering
        \footnotesize
        \def\tableheight{.135}
        \setlength{\tabcolsep}{0.3mm}
        \begin{tabular}{ccccc}
            \toprule
            \multicolumn{2}{c}{\textbf{Input}} & \multicolumn{3}{c}{\textbf{Scene Representation}} \\
            \cmidrule(lr){1-2} \cmidrule(lr){3-5} 
            \multicolumn{2}{c}{GT Depth $\longrightarrow$ Pseudo Depth}
            & Grids &  Points & Gaussians  \\ 
            \includegraphics[height=\tableheight\linewidth]{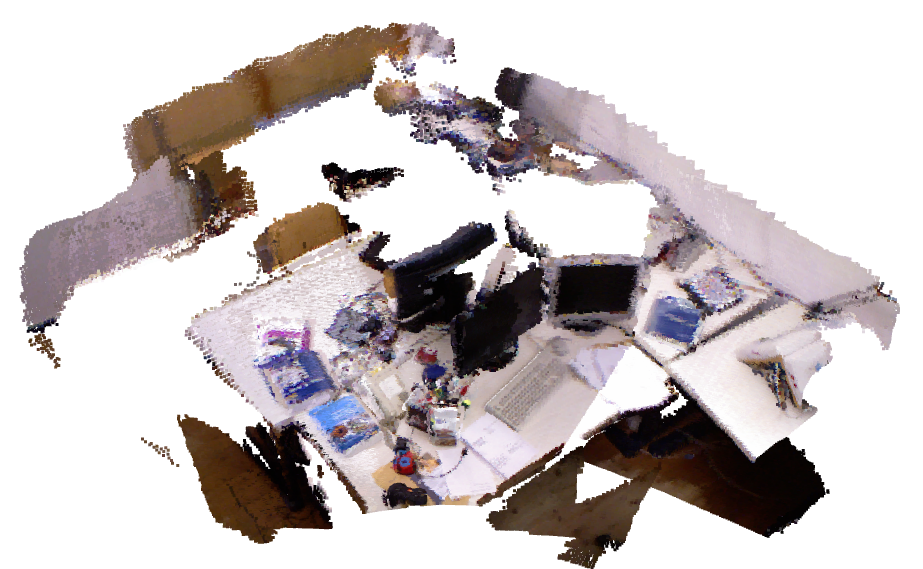} 
            &\includegraphics[height=\tableheight\linewidth]{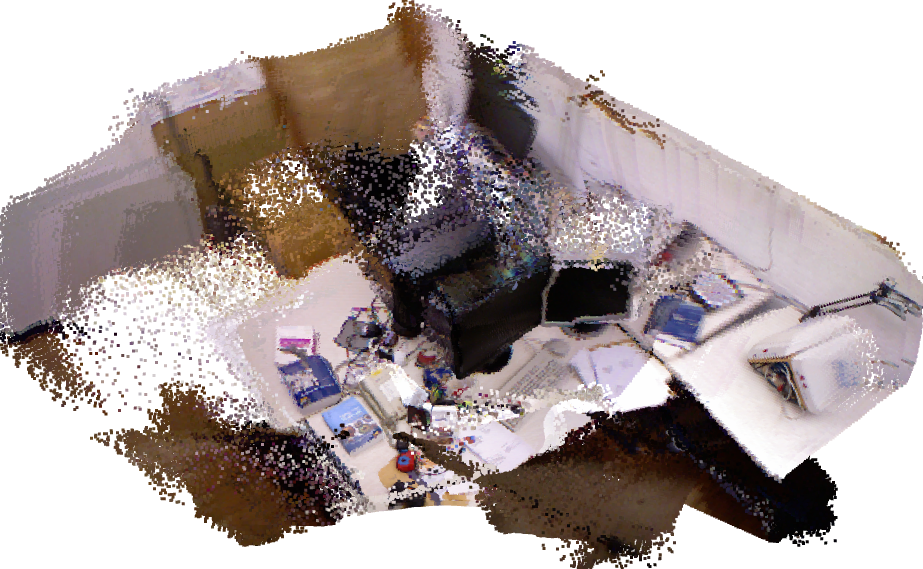}  &\includegraphics[height=\tableheight\linewidth]{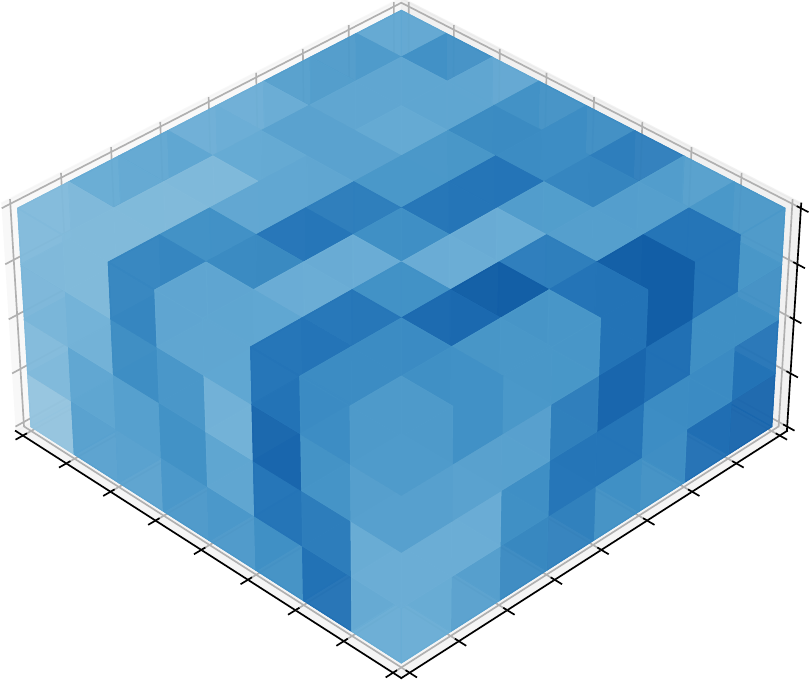}  & 
            \includegraphics[height=\tableheight\linewidth]{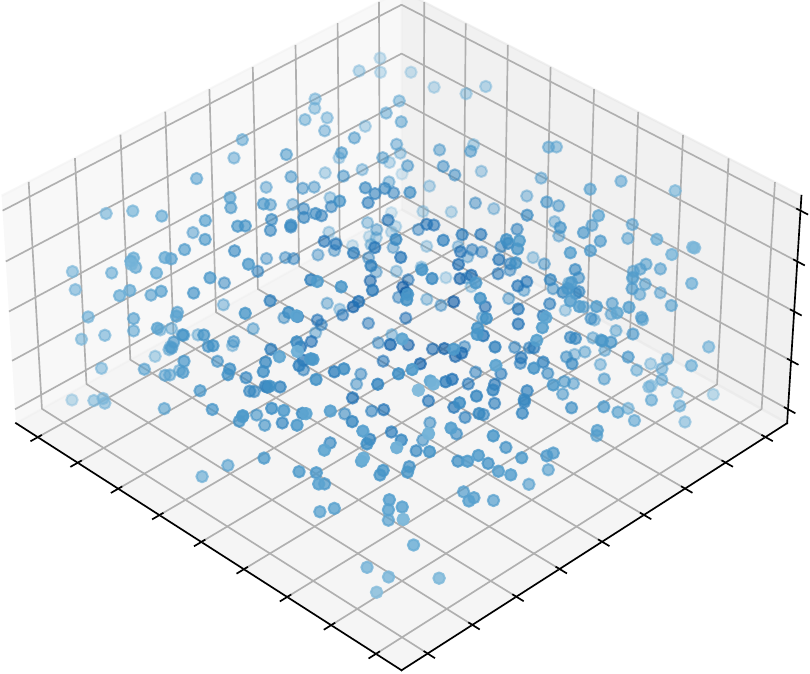} 
            & \includegraphics[height=\tableheight\linewidth]{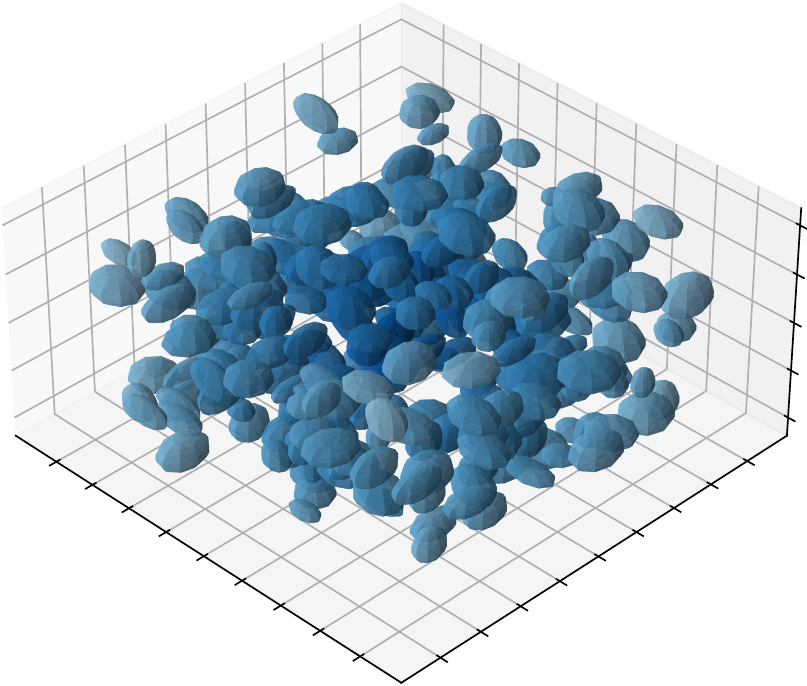}  \\ \midrule 
            \multicolumn{2}{c}{Decoding Strategy}  & MLP & MLP & Splatting   \\
            \multicolumn{2}{c}{Photometric Loss} & Sparse & Sparse & Dense
             \\
            \multicolumn{2}{c}{Differentiable Rendering}  & \cmark & \cmark & \cmark   \\
            \multicolumn{2}{c}{Data-adaptive Shapes}  & \xmark & \cmark & \cmark   \\
            \multicolumn{2}{c}{Optimizable Attributes}
             & \xmark & \xmark & \cmark   \\
            \bottomrule
        \end{tabular}
        \vspace{3mm}
    \end{minipage}
}
\subfigure[Scene Representations by GT Depth (top) and Pseudo Depth (bottom)]{
    \begin{minipage}{0.48\textwidth}
        \centering
        \small
        \def\myheight{.2}
        \setlength{\tabcolsep}{0.3mm}
        \begin{tabular}{ccc}
            Grids (Mesh) & Points & Gaussians \\
            \includegraphics[height=\myheight\linewidth]{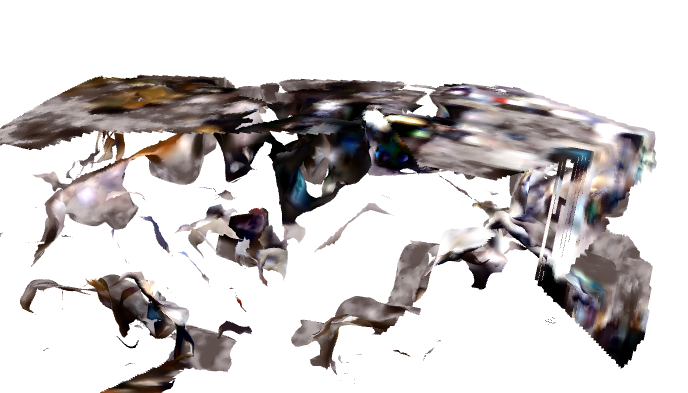} &
            \includegraphics[height=\myheight\linewidth]{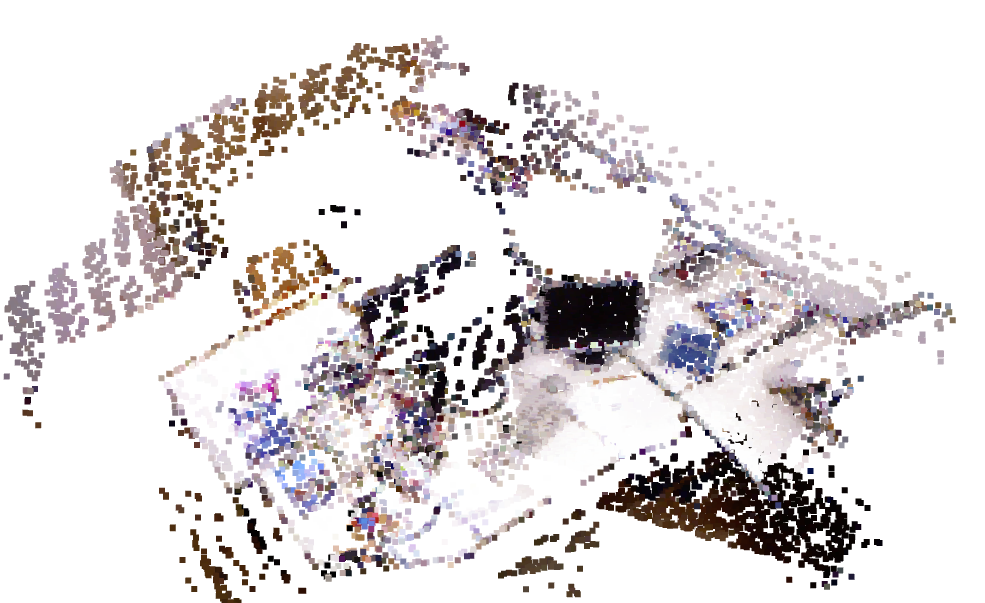} &
            \includegraphics[height=\myheight\linewidth]{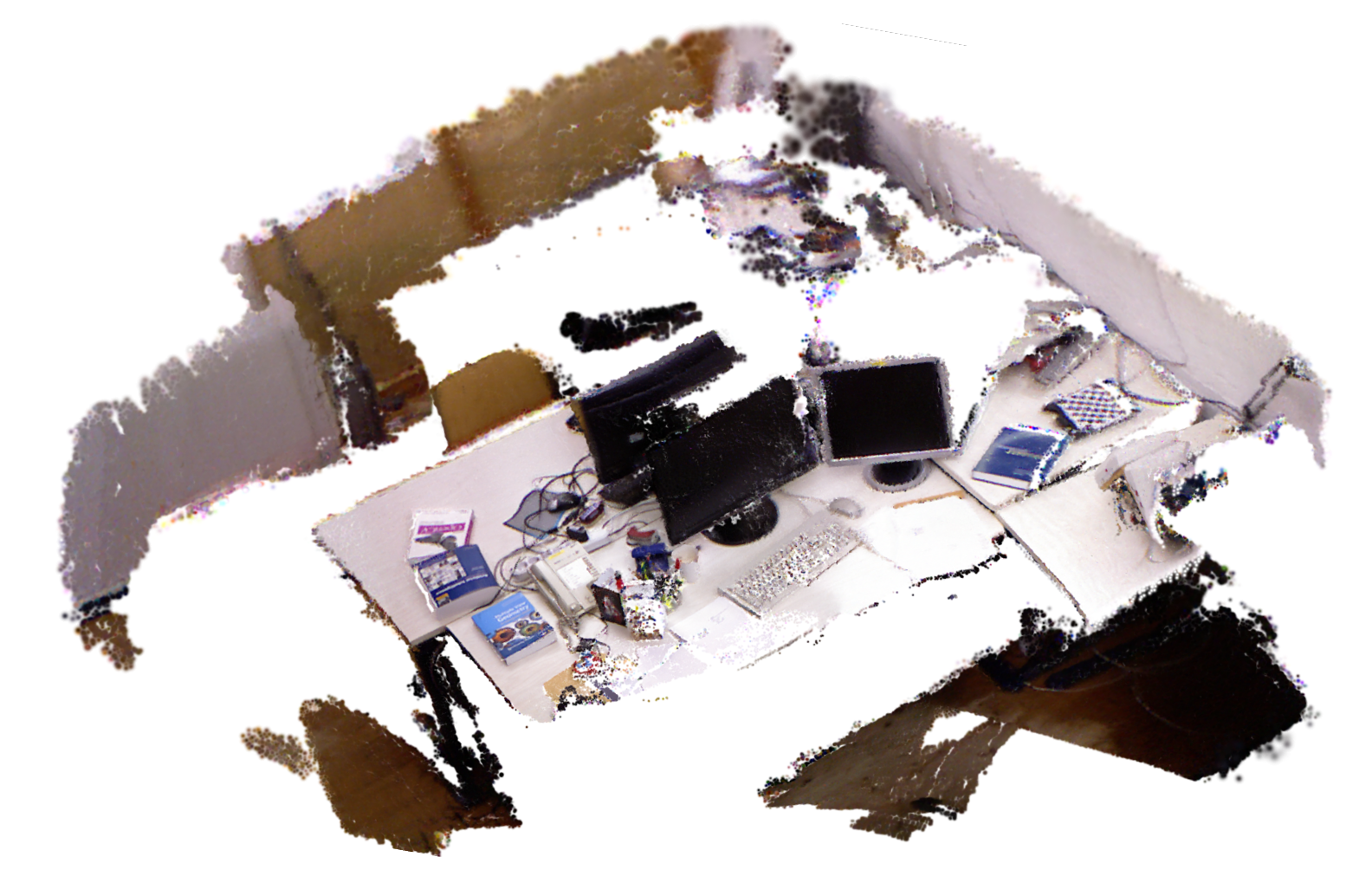}\\
            \includegraphics[height=\myheight\linewidth]{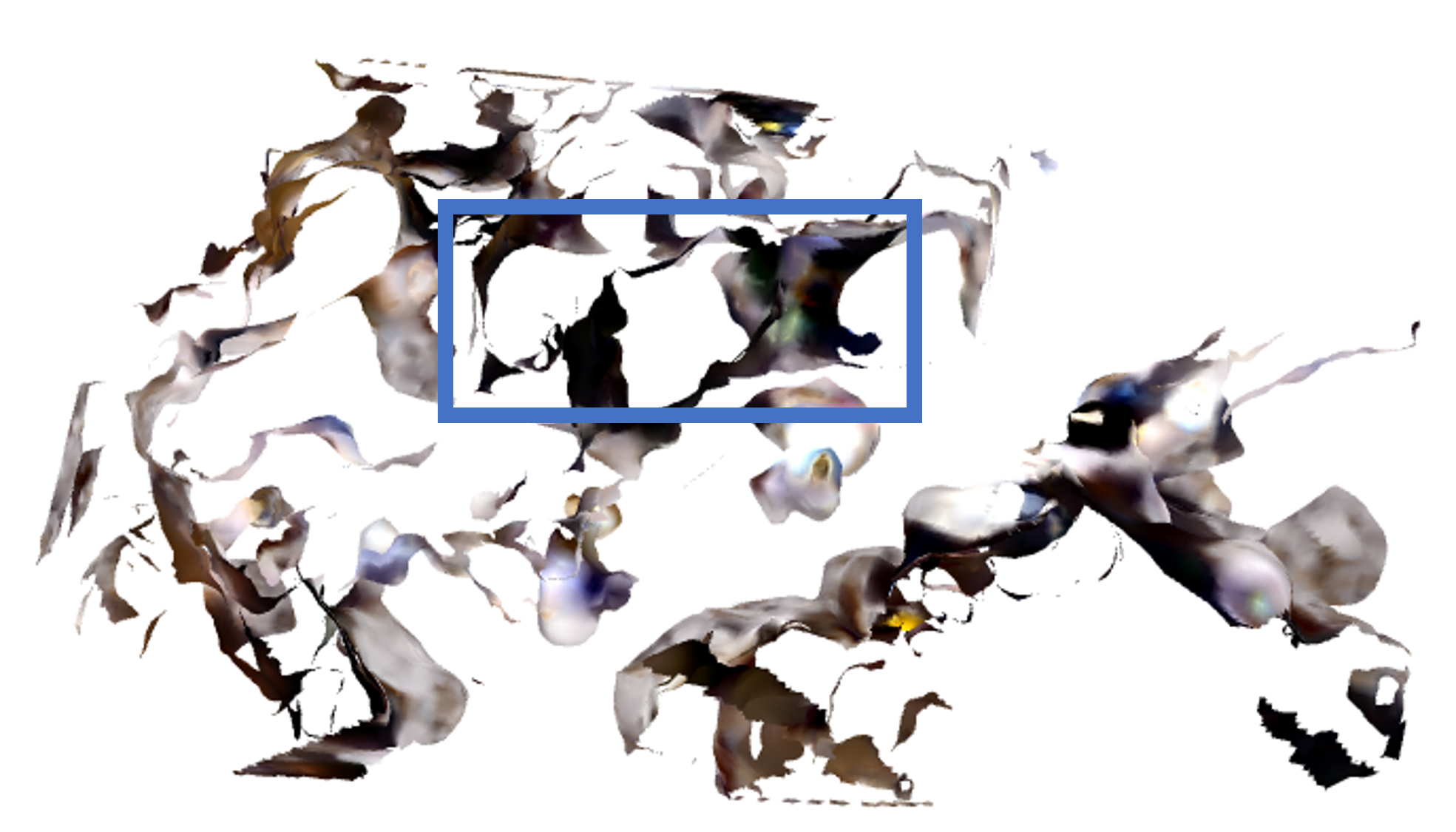} &
            \includegraphics[height=\myheight\linewidth]{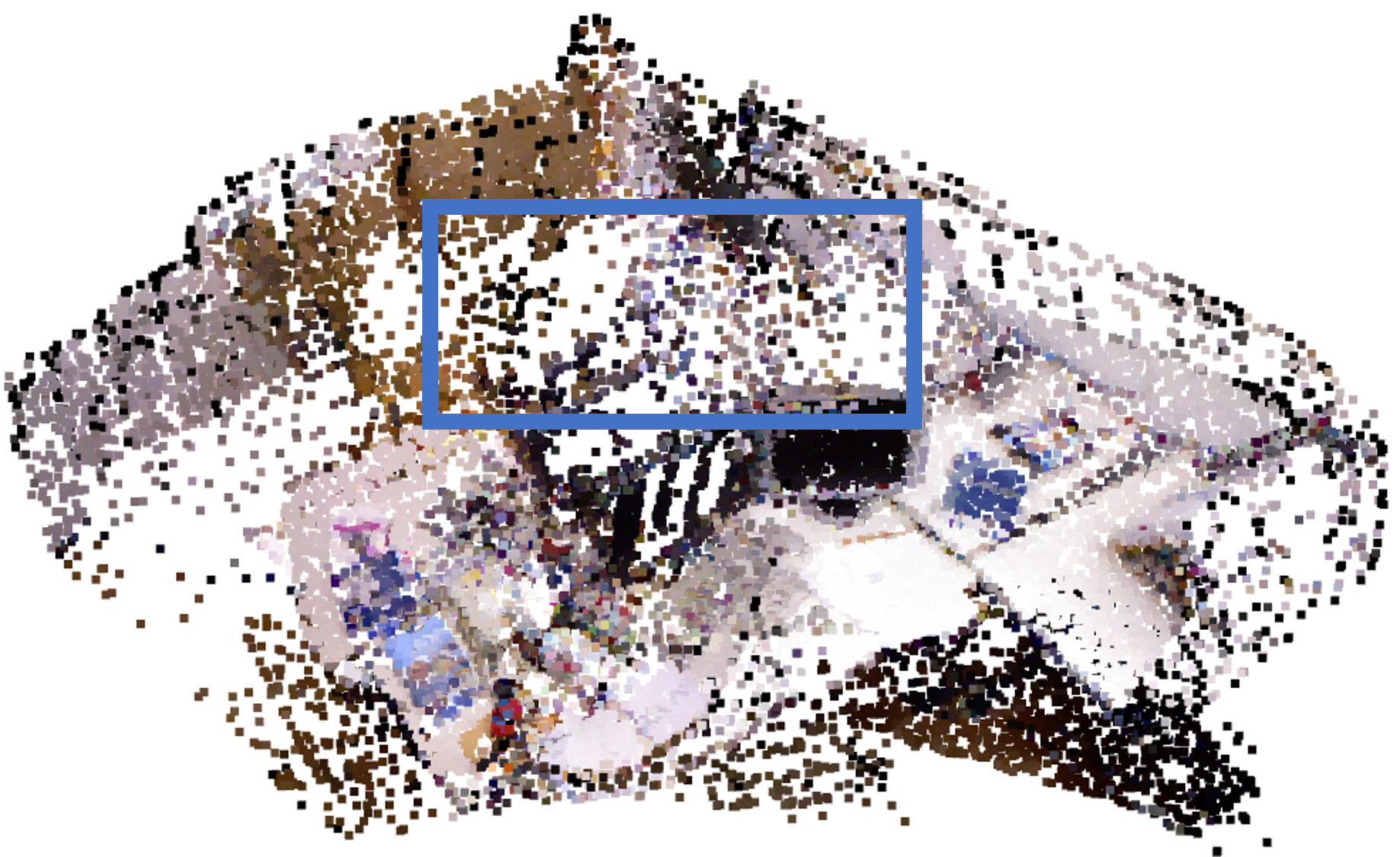} &
            \includegraphics[height=\myheight\linewidth]{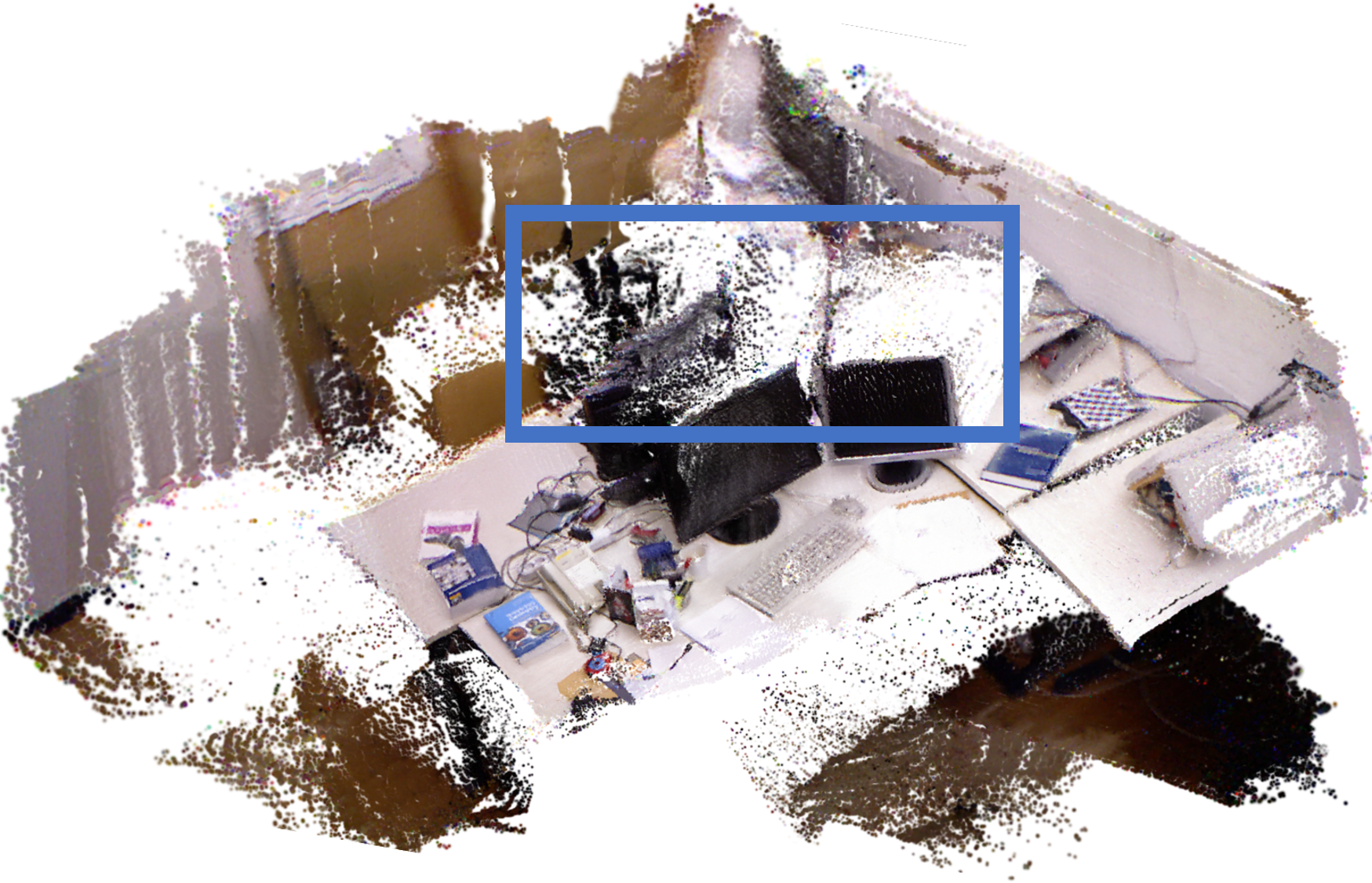} \\
        \end{tabular}
        \vspace{3mm}
    \end{minipage}
}
\vspace{-3mm}
\caption{Comparison of different scene representations. (a) We compare Grids, Points, and Gaussians on GT depth and pseudo depth, highlighting differences in decoding strategies, photometric loss types, differentiable rendering, data-adaptive shapes, and optimizable attributes. (b) The Gaussian representation demonstrates a high level of adaptability when applied to pseudo-depth, as it naturally accommodates depth uncertainties, leading to a more stable and realistic 3D reconstruction.}
\vspace{-7mm}
\label{fig:toy_experiment}
\end{figure*}

\section{Our Approach}

In this section, we first compare various mapping representations using both ground truth (GT) and pseudo-depth inputs. Next, we introduce an effective baseline for online RGB SLAM without GT depth as input. Finally, we propose a robust and efficient graph rendering method for fast, feed-forward pose tracking.

\subsection{Depth Estimation-based 3D Gaussian Mapping}
In this work, we address the practical task of online 3D reconstruction. Given a pose-free streaming RGB video input $\{I_1, I_2, \dots, I_t\}$ with known camera intrinsics $M$, the method incrementally reconstructs the 3D scene. At each time instant $t$, the output $P_t$ consists of the point cloud corresponding to the current image $I_t$ in world coordinates.

Balancing accuracy and efficiency in online 3D reconstruction is a challenging problem. Unlike offline methods, where all RGB images are available for simultaneous processing, the observations in online 3D reconstruction are typically partial at any given time. If the full set of previously observed images (${I_1, I_2, \dots, I_t}$) is used at each time step $t$, the computational cost increases continuously, resulting in $\mathcal{O}(t^2)$ time complexity. On the other hand, if $P_t$ is predicted solely from the current image $I_t$ or a few nearest frames, there may not be enough information to accurately reconstruct the 3D geometry or align it with the world coordinate system, given the limited RGB observations. To address this, Spann3R~\cite{wang20243d} introduces a spatial memory to store both long-term and short-term features extracted from previous frames. At each time step, the current frame $I_t$ interacts with this memory to aggregate past geometric information, which is then processed by an end-to-end prediction module to directly output the 3D point clouds. While this approach is efficient, the implicit feature-based memory suffers from catastrophic forgetting, causing the quality of $P_t$ to degrade over time.

To address this challenge, we propose an explicit solution by leveraging a visual SLAM method to estimate the camera pose $G_t$ in the world coordinate system at each time step. By tracking the camera movement, we can decompose the online 3D reconstruction task into two subproblems: online camera pose tracking and monocular depth estimation. 
However, state-of-the-art rendering-based SLAM methods~\cite{keetha2024splatam,sandstrom2023point,zhu2022nice} typically require RGB-D images as input, which places high demands on the sensors. To alleviate this, we reduce the data requirements by using a monocular depth estimation model to predict pseudo-depth maps from RGB images. This approach allows us to perform 3D reconstruction without the need for specialized RGB-D sensors. While pseudo-depth maps provide a convenient alternative, they are often prone to errors, particularly in the presence of occlusions, textureless regions, and depth ambiguities. As a result, pseudo-depth is generally less accurate than ground-truth depth, leading to potential inaccuracies in the 3D reconstruction. To overcome these challenges, it is crucial to select an appropriate mapping representation that can effectively handle these errors and maintain robust reconstruction performance despite the imperfections in the pseudo-depth data.

\begin{figure*}[t]
\centering
\def\myheight{1.9in}
\subfigure[Mapping]{
	\includegraphics[height=\myheight]{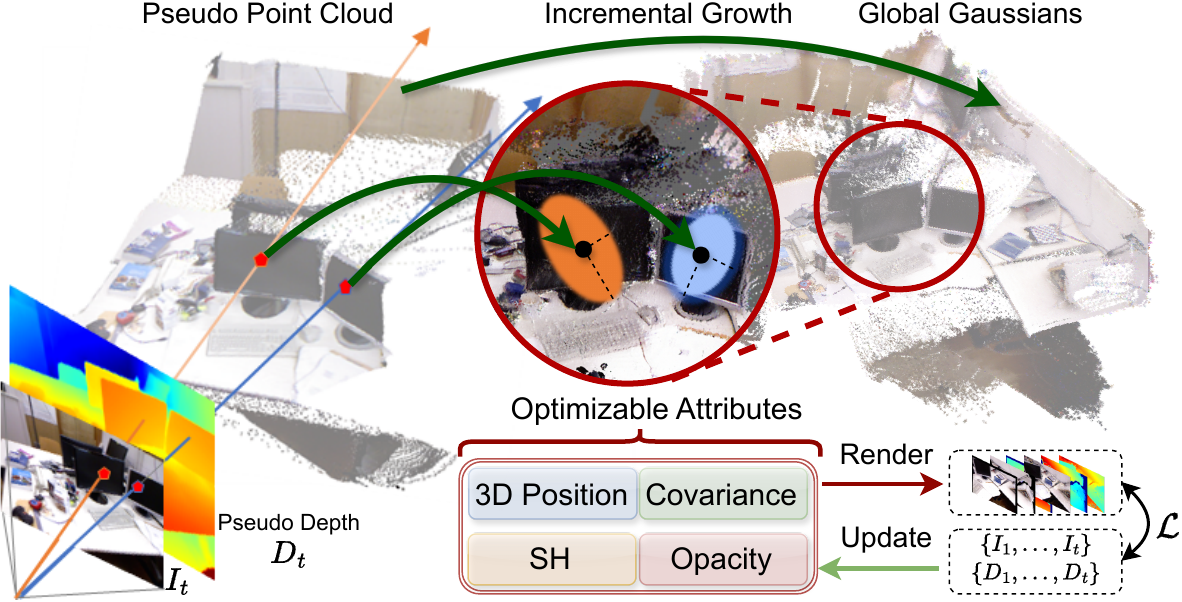}
}
\subfigure[Tracking]{
	\includegraphics[height=\myheight]{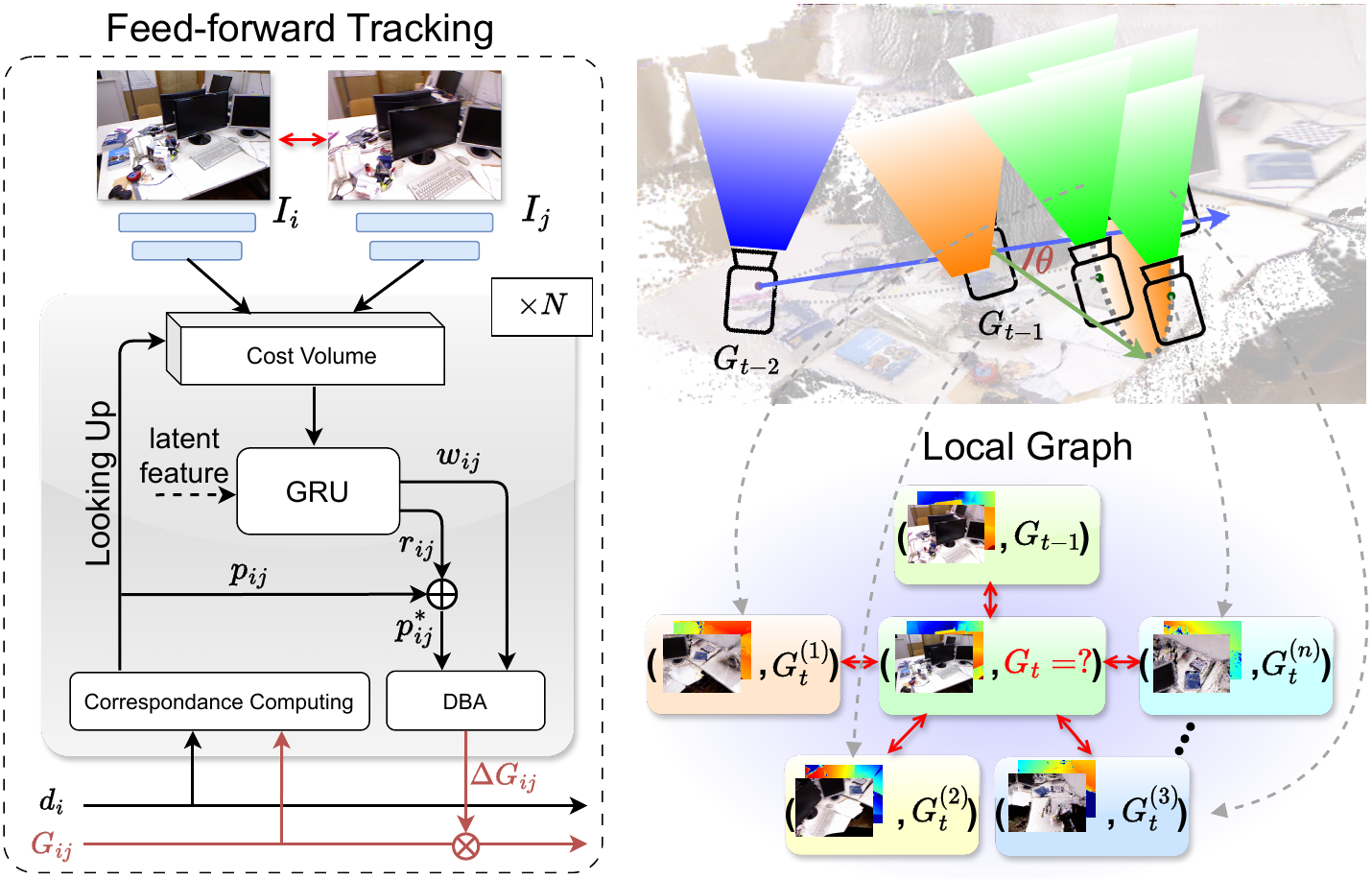}
}
\vspace{-3mm}
\caption{Overview of the proposed online 3D reconstruction pipeline. The pipeline consists of two main components: 1) \textbf{Gaussian Growth}, where the scene representation is incrementally updated by refining the 3D Gaussians using pseudo-depth data, 2) \textbf{Feed-forward Camera Tracking}, where we directly estimate relative poses between RGB frames, and further improving the estimation through local graph rendering, significantly enhancing system efficiency.
}
\vspace{-7mm}
\label{fig:pipeline}
\end{figure*}

\noindent \textbf{Adaptability to Pseudo-Depth: Grids, Points, and Gaussians.} Fig.~\ref{fig:toy_experiment} illustrates the key differences and qualitative performance of grids~\cite{zhu2022nice}, points~\cite{sandstrom2023point}, and Gaussians~\cite{keetha2024splatam,matsuki2024gaussian} in handling scene reconstruction when ground-truth (GT) depth is replaced by pseudo-depth inputs.
\textbf{Grid-based representations} partition the scene into a regular 3D grid, where each cell corresponds to a fixed volume of space. The unit element of the grid can be represented as $G_{i,j,k} = \{x, y, z\} \in \mathbb{R}^3$, where $(i, j, k)$ are the indices of the grid cell in a 3D array. When using pseudo-depth, inaccuracies can cause misalignment and discontinuities in the grid structure, as small errors in depth estimation lead to significant misplacements of grid cells. 
\textbf{Point-based representations} offer greater flexibility and memory efficiency by storing only sparse 3D points, wherer a point can be represented as $p = \{x, y, z\} \in \mathbb{R}^3$. Nevertheless, in the absence of smoothing or probabilistic encoding, errors in pseudo-depth are directly manifested in the representation, resulting in inconsistencies and gaps in the reconstructed scene.
In contrast, the unit element of \textbf{Gaussian-based representations} can be represented as $f(\mathbf{x}) = o\cdot \exp\left(-\frac{\|\mathbf{x} - \boldsymbol{\mu}\|^2}{2r^2}\right)$, where $\mathbf{x}$ is the 3D position, $\boldsymbol{\mu} \in \mathbb{R}^3$ is its center position, $r$ controls the spread or radius, and $o \in [0,1]$ denotes the opacity. It naturally encodes uncertainty, allowing for smoother transitions between neighboring regions.  In addition, the geometric structure of the 3D Gaussians is optimizable, meaning the model can adapt and refine the scene representation based on the observed data. When applied to pseudo-depth, the Gaussian representation "blurs" depth inaccuracies by spreading the uncertainty across a region, resulting in more continuous and stable 3D reconstruction.

\noindent \textbf{Baseline.} Building on the adaptability of Gaussian representations to pseudo-depth inputs, we propose a simple online 3D reconstruction baseline that incrementally grows the scene representation using pseudo-depth data, as shown in Fig.~\ref{fig:pipeline} (a). The optimization alternates between refining the Gaussian representation and the camera poses. 
Given an initial Gaussian representation $\mathcal{G}_{t-1}$ from the previous frame and the current camera pose $G_i$, we perform the following alternating optimization steps:

\begin{itemize}
    \item \textbf{Gaussian Optimization:}  
    With fixed poses, we minimize the difference between the rendered images (and depth maps) and their corresponding input values:
\begin{equation} \small
\begin{aligned}
        \mathcal{G}_t = \text{min}_{\mathcal{G}} &\sum_{i \in \mathcal{N}(t)} (\left| \mathcal{R}_{img}(\mathcal{G}_{t-1} \cup \mathcal{P}(D_t), G_i) - I_i \right| \\
        &~~~~~~+ \left| \mathcal{R}_{dep}(\mathcal{G}_{t-1} \cup \mathcal{P}(D_t), G_i) - D_i \right|),
\end{aligned}
\end{equation}
    \item \textbf{Pose Optimization:}  
    With fixed 3D Gaussian representations, we optimize the camera pose to minimize the discrepancy between the rendered images (and depth maps) and input values:
    \begin{equation} \small
    \begin{aligned}
            G_t &= \text{min}_{G} (\left| \mathcal{R}_{img}(\mathcal{G}_{t-1}, G) - I_t \right| \\
            &~~~~~~~~~~~+ \left| \mathcal{R}_{dep}(\mathcal{G}_{t-1}, G) - D_t \right|),
    \end{aligned}
    \end{equation}
\end{itemize}
where $\mathcal{R}_{img}$ and $\mathcal{R}_{dep}$ represent the rendering image and depth operation, respectively, $\mathcal{G}$ denotes the full set of 3D Gaussians, and $\mathcal{N}(t)$ is the set of recent timestamps, including $t$. The function $\mathcal{P}$ is a projection and prediction module that initializes the 3D Gaussians based on the pseudo depth $D_t$ along with additional parameters $M$ and $G_t$. This alternating optimization approach continuously refines both the Gaussian representation and the camera pose, making it well-suited for real-time scene reconstruction with pseudo-depth inputs.

\subsection{Graph Rendering for Feed-forward Tracking}
Although leveraging 3D Gaussians for scene mapping can significantly mitigate the impact of inaccurate depth estimation, we identify the camera pose tracking operation as the primary efficiency bottleneck in the overall 3D reconstruction system. Currently, the tracking stage relies on iterative optimization of pose parameters, which consumes a substantial portion of the inference time. To overcome this, we propose substituting the optimization-based pose tracking with a feed-forward prediction method. Specifically, we introduce a learnable module that directly predicts the relative pose between two RGB frames. Given that neural network inference is much faster than performing hundreds of backpropagation steps through the 3D Gaussian rendering, we believe that using feed-forward prediction will significantly accelerate pose tracking, leading to a substantial improvement in overall system efficiency.

\noindent \textbf{Feed-forward Pose Prediction.} Inspired by DROID-SLAM~\cite{teed2021droid}, we propose predicting the relative pose between two frames,
say $I_{i}$ and $I_{j}$ for simplicity, 
by leveraging geometric constraints between the camera poses and dense per-pixel depth, as shown in Fig.~\ref{fig:pipeline} (b). Specifically, we solve for the relative pose by maximizing the alignment between these constraints and the current optical flow estimate, which is provided by a recurrent architecture such as RAFT~\cite{teed2020raft}.
At the start of each iteration, we use the current estimates of poses $G$ and pseudo-depth maps $D$ to compute the correspondences. Given a pixel coordinate, $p_i \in \mathbb{R}^{H \times W \times 2}$ in frame $I_i$, we compute the dense correspondence field $p_{ij} \in \mathbb{R}^{H\times W \times 2}$ by:
\begin{equation}
    p_{ij} = \Pi_c(G_{ij} \circ \Pi_c^{-1}(p_i, d_i)), 
    ~~~ G_{ij} = G_j \circ G_i^{-1},
\end{equation}
where $\Pi_c$ represents the camera model that projects a set of 3D points onto the image plane, while $\Pi_c^{-1}$ is the back-projection function, which maps the depth value $d$ and pixel coordinates $p_i$ to a 3D point cloud. $p_{ij}$ denotes the coordinates of pixels $p_i$ mapped into frame $j$ using the estimated pose and depth.
The correspondence field $p_{ij}$ provides the initial dense matching between frames, which is updated by the GRU~\cite{chung2014empirical} to refine the flow field $\mathbf{r}_{ij}$ and confidence map $\mathbf{w}_{ij}$. The GRU uses both the correlation features and flow features to predict these updates, which are then used to improve the pixel-wise correspondence. The Gauss-Newton algorithm is adopted to take the corrected correspondences $p_{ij}^* = \mathbf{r}_{ij} + p_{ij}$ and compute the residuals of camera poses $\Delta G_{ij}$ while keeping the pseudo depths ${d}$ fixed, by minimizing the re-projection error:
\begin{equation} \label{equ:DBA}
    \Delta G_{ij} = \text{min}_{G} ||{p_{ij}^* - \Pi_c(G_{ij} \circ \Pi_c^{-1}(p_i, d_i)) ||}^2.
\end{equation}
The mathematical solution is implemented as CUDA operations to enable efficient end-to-end inference. To achieve this, we adopt the DBA layer proposed in \cite{teed2021droid}, which is optimized for high-performance execution on GPUs.

\begin{table}[t]
    \footnotesize
    \centering
    \renewcommand{\arraystretch}{0.8}
    \caption{\textbf{Tracking Performance on Replica~\cite{straub2019replica}} (ATE RMSE $\downarrow$ [cm]). }
    \label{fig:figure1}
    \def\myheight{.15}
    \def\mywidth{.24}
    \setlength{\tabcolsep}{1pt}
    \begin{tabular}{cccc}
        \toprule
        \multirow{2}{*}{SplaTAM~\cite{keetha2024splatam}} & Ours  & \multirow{2}{*}{Ours} & \multirow{2}{*}{Ground Truth} \\
        & (w/o LGR) &  & \\
        \midrule
        \blurowcolor
        \multicolumn{4}{c}{Trajectory of \texttt{Office 0 (Of0)}} \\
        \blurowcolor
        \includegraphics[width=\mywidth\linewidth,height=\myheight\linewidth]{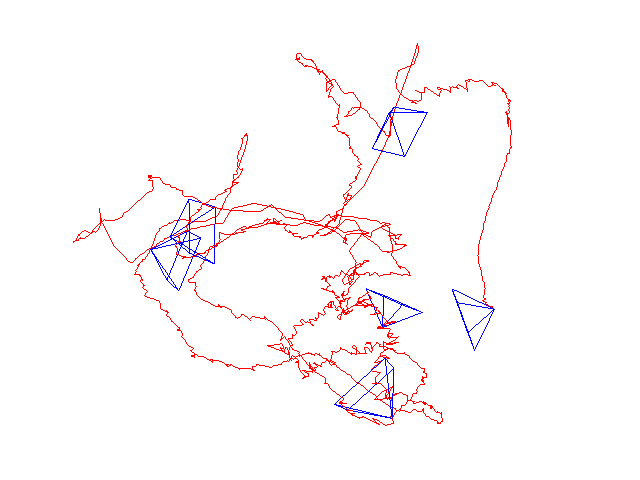}&
    	\includegraphics[width=\mywidth\linewidth,height=\myheight\linewidth]{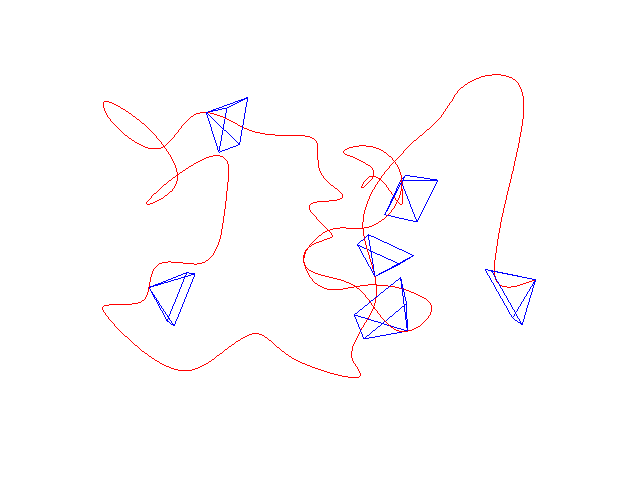}&
    	\includegraphics[width=\mywidth\linewidth,height=\myheight\linewidth]{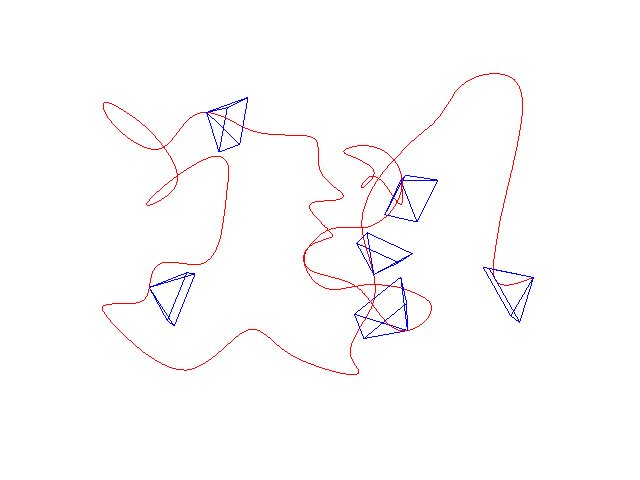}&
    	\includegraphics[width=\mywidth\linewidth,height=\myheight\linewidth]{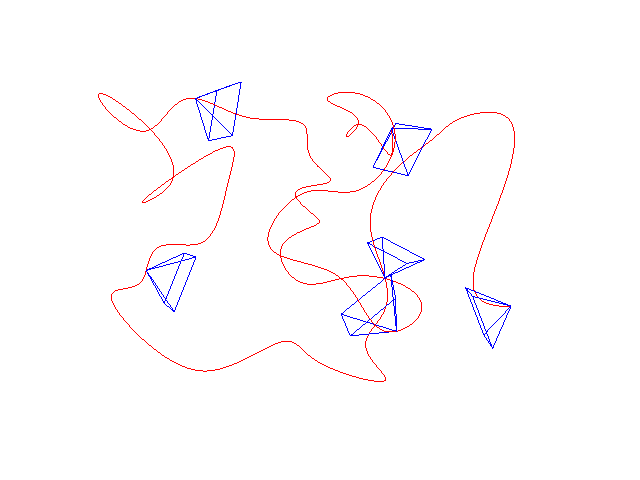} 
    \\
   \blurowcolor RMSE: 43.7 & RMSE: 21.6  & RMSE: 22.0 &  \\ \midrule
    	\blurowcolor \multicolumn{4}{c}{Trajectory of \texttt{Office 1 (Of1)}} \\
        \blurowcolor \includegraphics[width=\mywidth\linewidth,height=\myheight\linewidth]{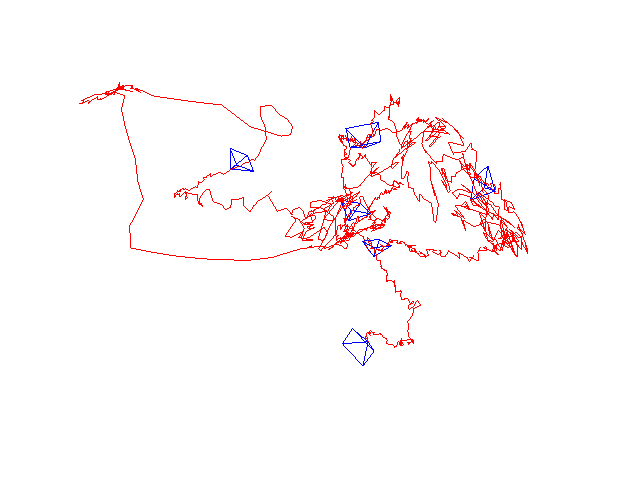}&
    	\includegraphics[width=\mywidth\linewidth,height=\myheight\linewidth]{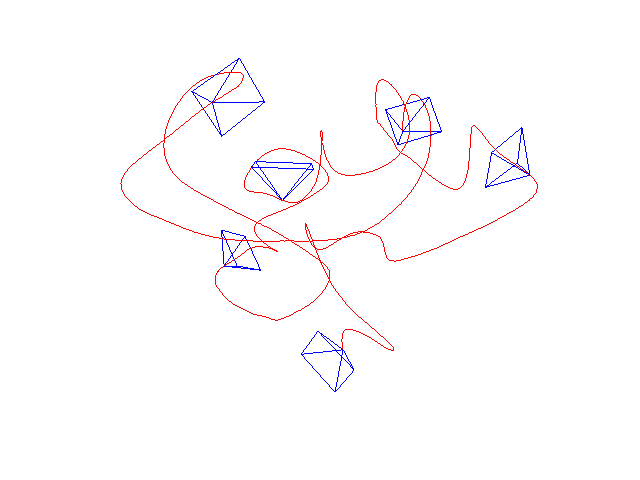}&
    	\includegraphics[width=\mywidth\linewidth,height=\myheight\linewidth]{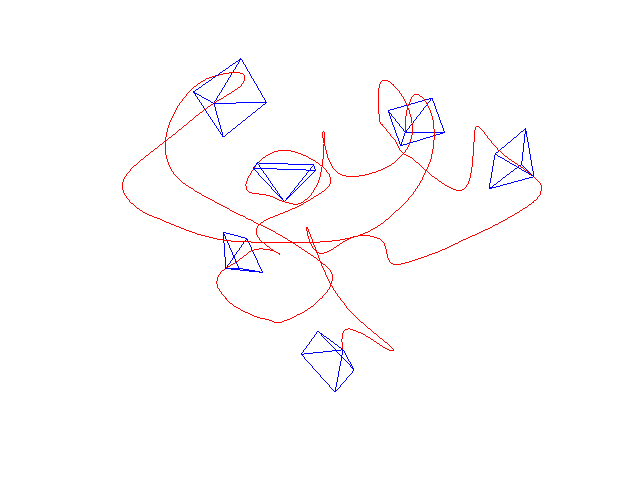}&
    	\includegraphics[width=\mywidth\linewidth,height=\myheight\linewidth]{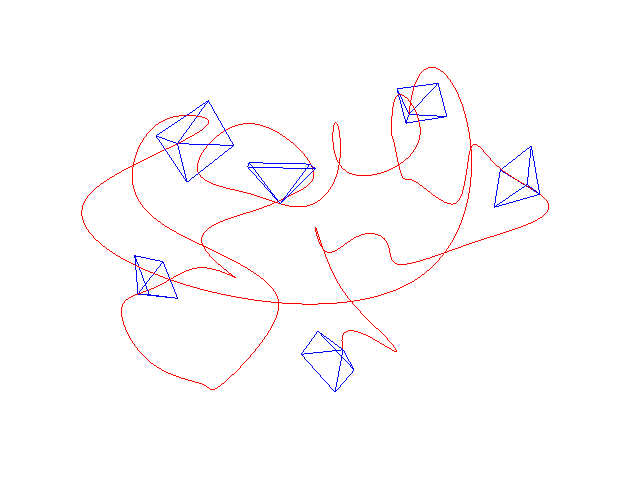} 
    \\
   \blurowcolor RMSE: 48.5 & RMSE: 11.3  & RMSE: 11.0 &  \\ \midrule
        \blurowcolor \multicolumn{4}{c}{Trajectory of \texttt{Room 1 (R1)}} \\
       	\blurowcolor \includegraphics[width=\mywidth\linewidth,height=\myheight\linewidth]{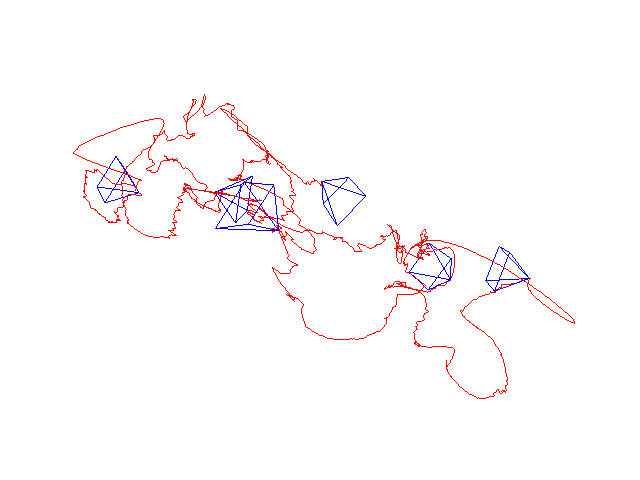}&
    	\includegraphics[width=\mywidth\linewidth,height=\myheight\linewidth]{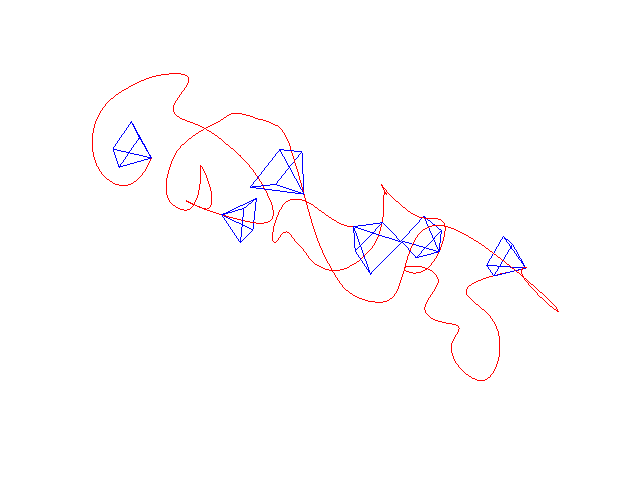}&
    	\includegraphics[width=\mywidth\linewidth,height=\myheight\linewidth]{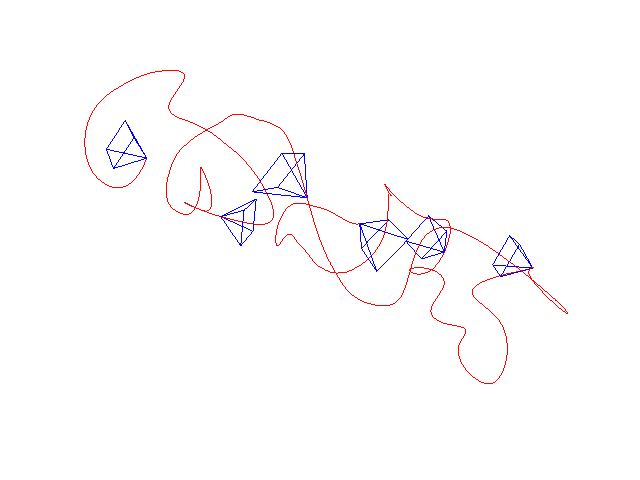}&
    	\includegraphics[width=\mywidth\linewidth,height=\myheight\linewidth]{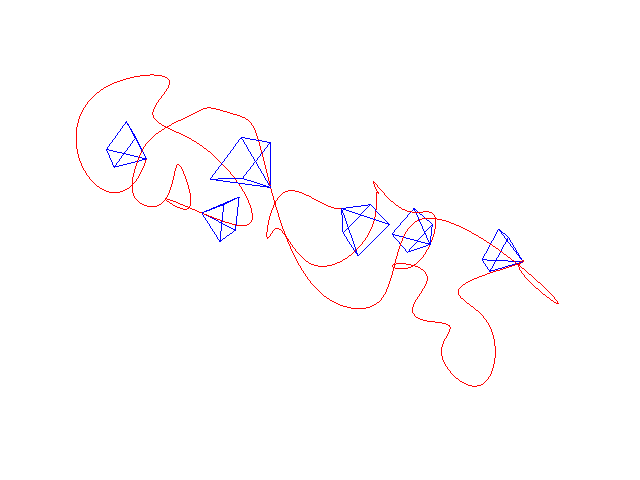} 
    \\
       \blurowcolor RMSE: 37.4 & RMSE: 15.2  & RMSE: 10.5 &  \\ \midrule
        \blurowcolor \multicolumn{4}{c}{Average RMSE of \texttt{\{R0, R1, Of0, Of1, Of2\}}} \\ [5pt]  
        \blurowcolor      31.19 & 16.12 & \textbf{15.49} &  \\
       \bottomrule
    \end{tabular}
		\vspace{0.05cm}
		
    \vspace{-4mm}
\end{table}

\noindent \textbf{Local Graph Rendering (LGR).} The 2-frame feed-forward pose prediction offers significantly faster pose tracking than traditional optimization-based methods (see Table~\ref{tab:runtime}). However, predicting the accurate relative pose from just two RGB images is often unreliable due to challenges such as changing illumination, motion blur, and textureless regions, all of which make optical flow estimation difficult.

To address this issue, we construct a local graph using multiple image pairs to predict the relative pose. Since distant images typically have limited overlap with the target frame \(I_t\), we focus on building a \emph{local} graph that includes several image pairs with better overlap, thereby enabling more reliable optical flow estimation. We construct a local graph \( \mathcal{E} \) centered around the target frame \( I_t \), consisting of \( n \) edges, as shown in Fig.~\ref{fig:pipeline} (b). These edges establish correspondences between image pairs with sufficient overlap, enabling more reliable optical flow estimation and more accurate relative pose prediction. Specifically, the local graph is defined as follows:
\begin{itemize}
\item \textbf{Nodes:} Each node \( I_{t}^{(k)} \) represents a rendered frame with a sampled pose \( G_{t}^{(k)} \), generated by splatting the Gaussian representation \( \mathcal{G}_{t} \) using \( G_{t}^{(k)} \):
\begin{equation}
    I_{t}^{(k)} = \mathcal{R}_{img}(\mathcal{G}_{t}, G_{t}^{(k)}), ~~D_{t}^{(k)} = \mathcal{R}_{dep}(\mathcal{G}_{t}, G_{t}^{(k)}).
\end{equation}
\item \textbf{Edges:} One edge connects \( I_t \) to the previous frame \( I_{t-1} \), and the remaining \( n-1 \) edges connect \( I_t \) to nearby rendered images \( I_t^{(k)} \), where \( k \in \{1, 2, \dots, n-1\} \). Each edge represents an image pair \( (I_t, {t-1}) \) or \( (I_t, I_t^{(k)}) \) whose relative pose is known.
\end{itemize}
The relative pose prediction is then optimized by minimizing the reprojected error across all image pairs in the local graph, as described in Equation~\ref{equ:DBA}. 

\begin{table}[t]
\label{tab:tum_tracking1}
\label{tab:tum_tracking2}
\setlength{\tabcolsep}{0.004\linewidth}
\footnotesize
\caption{\textbf{Tracking Performance on TUM-RGBD~\cite{Sturm2012ASystems}} (ATE RMSE $\downarrow$ [cm]). The upper part shows the results of each method when the input is GT depth, while the lower part shows the results when the input is pseudo depth.}
\centering
\renewcommand{\arraystretch}{1.1}
\begin{tabular}{cccccc}
\toprule
{\textbf{Method}} & {\textbf{Depth}}& {\textbf{Average}} & \texttt{fr1/desk} &  \texttt{fr1/desk2}  & {\textbf{Time/Frame}}\\
\midrule
\grayrowcolor Kintinuous \cite{whelan2012kintinuous} & GT & 5.40 & 3.70 & 7.10  & -\\ [0.8pt]  \noalign{\vskip 1pt}
\grayrowcolor ElasticFusion \cite{whelan2015elasticfusion} & GT & 4.68 & 2.53 & 6.83  & - \\ [0.8pt]  \noalign{\vskip 1pt}
\grayrowcolor ORB-SLAM2 \cite{ORB-SLAM2} & GT & 1.90 & 1.60 & 2.20 & - \\ [0.8pt]  \noalign{\vskip 1pt}
\hdashline
\grayrowcolor NICE-SLAM \cite{zhu2022nice} & GT& 4.63 & 4.26 & 4.99 &  3.67 s\\ [0.8pt]  \noalign{\vskip 1pt}
\grayrowcolor Vox-Fusion \cite{yang2022vox}& GT & 4.76 & 3.52 & 6.00 & 5.21 s \\  [0.8pt]  \noalign{\vskip 1pt}
\grayrowcolor Point-SLAM \cite{sandstrom2023point} & GT & 4.44 & 4.34 & 4.54 & 3.79 s  \\ [0.8pt]  \noalign{\vskip 1pt}
\grayrowcolor GS-SLAM \cite{yan2024gs} & GT & -  &  3.30 & - & - \\ [0.8pt]  \noalign{\vskip 1pt}
\grayrowcolor GS-ICP \cite{ha2025rgbd} & GT & - & 2.70 & - & - \\ [0.8pt]  \noalign{\vskip 1pt}
\grayrowcolor SplaTAM \cite{keetha2024splatam}& GT & 4.95 & 3.35 & 6.54 & 2.61 s \\ [0.8pt]  \noalign{\vskip 1pt}
\midrule
SplaTAM \cite{keetha2024splatam} & Pseudo &  8.89 & \textbf{7.92} &	9.87  & 2.61 s\\ [0.8pt]  \noalign{\vskip 1pt}
Ours (w/o LGR) & Pseudo & 12.33 & 17.01	 & 7.66	 & \textbf{0.19 s} \\ [0.8pt]  \noalign{\vskip 1pt}
Ours  & Pseudo & \textbf{8.62} & 9.67  &	\textbf{7.58} &  0.53 s\\ \bottomrule
\vspace{-7mm}
\end{tabular}
\end{table}

\noindent \textbf{Spherical Pose Sampling.} 
Since the exact pose \( G_t \) corresponding to \( I_t \) is unknown, and in order to sample poses with the maximum overlap with \( I_t \) for rendering images and depth maps, we propose a spherical sampling technique based on camera inertia. First, we generate an approximate absolute pose \( \hat{G}_t \) of \( I_t \) by leveraging the camera's inertia on the translation part, which is computed as:
\begin{equation}
\hat{G}_t = \left[ \mathbf{R}_{t-1} \right] \left[2 \mathbf{T}_{t-1} - \mathbf{T}_{t-2}\right],
\end{equation}
where \( \mathbf{T} \) represents the translation component of the pose \( G \), and the rotation part \( \mathbf{R} \) remains unchanged.

Next, to sample \( n-1 \) camera poses around \( \hat{G}_t \), we perform spherical sampling on the translation component of the pose. Specifically, the sampled camera poses \( G_t^{(k)} \) are given by:
\begin{equation} \label{eq:pose-sample}
G_t^{(k)} = \left[ \mathbf{R}_{t-1} \right] \left[ \alpha \| \mathbf{T}_{t-1} - \mathbf{T}_{t-2} \| \cdot \hat{v}_i(\theta) \right],
\end{equation}
where \( \| \mathbf{T}_{t-1} - \mathbf{T}_{t-2} \| \) represents the magnitude (norm) of the camera's translation inertia, \(\alpha\) is a scaling factor, and \( \hat{v}_i(\theta) \) is a unit vector on the sphere offset by an angle \( \theta \) from the inertia direction. This ensures that the translation part of the camera poses is sampled along a spherical path, providing maximum overlap with the target frame \( I_t \), which enhances the robustness of pose prediction by utilizing multiple image pairs within the local graph.

\section{Experiment}
We first describe our experimental setup and then evaluate our method against state-of-the-art dense neural RGBD SLAM methods on Replica~\cite{straub2019replica} as well as the real-world TUM-RGBD~\cite{Sturm2012ASystems}.

\subsection{Experimental Settings}
\noindent \textbf{Implementation Details.}
On all datasets mentioned above, we conduct 3 parallel groups of experiment with different tracking methods: (1) iterative-Gaussian-rendering-based tracking; (2) feed-forward style tracking with no additional image rendering, i.e. we only adopt constraints provided by the frame pair \( (I_{t-1}, I_t) \); (3) feed-forward style tracking with \(N\) additional images rendered. All experiments use pseudo-depth inferred by the monocular depth model UniDepthV2~\cite{piccinelli2024unidepth}. We directly loaded the pre-trained model provided by DROID-SLAM to initialize the network parameters in the RAFT iteration module, including the GRU. Based on proper ablation study, we set \(N=6\), \(\alpha = 5.0\) and \(\theta = 30^\circ \) (see eq.\ref{eq:pose-sample}).

\noindent \textbf{Evaluation Metrics.}
We adopt the evaluation metrics for camera pose estimation and rendering performance as outlined in SplaTAM~\cite{keetha2024splatam}. Specifically, for RGB rendering performance, we evaluate using PSNR, SSIM, and LPIPS. For depth rendering, we use the Depth L1 loss. For camera pose estimation, we assess tracking accuracy using the average absolute trajectory error (ATE RMSE).

\noindent \textbf{Datasets.} 
The Replica dataset~\cite{straub2019replica} offers high-quality 3D reconstructions of indoor scenes with accurate depth maps and small camera pose displacements, making it ideal for evaluating scene understanding methods. In contrast, the TUM-RGBD dataset~\cite{Sturm2012ASystems} presents a more challenging benchmark, with sparse depth data, motion blur, and low-quality sensor information, making it a realistic test for handling real-world data.

\noindent \textbf{Baseline Methods.} 
The primary baseline method we compare against is SplaTAM~\cite{keetha2024splatam}, the previous state-of-the-art (SOTA) approach for dense render-based SLAM. Additionally, we compare with earlier dense SLAM methods such as Point-SLAM~\cite{sandstrom2023point}, NICE-SLAM~\cite{zhu2022nice}, Vox-Fusion~\cite{yang2022vox}, and ESLAM~\cite{johari2023eslam}. Finally, we also compare against traditional SLAM systems like Kintinuous~\cite{whelan2012kintinuous}, ElasticFusion~\cite{whelan2015elasticfusion}, and ORB-SLAM2~\cite{ORB-SLAM2} on TUM-RGBD, DROID-SLAM~\cite{teed2021droid} on the Replica dataset.

\subsection{Tracking Results}
As shown in Table~\ref{fig:figure1}, our method outperforms SplaTAM across various room configurations and scene offsets on the Replica dataset. This improvement is primarily due to the robust 3D Gaussian-based mapping, which enhances both tracking accuracy and stability. Notably, our method consistently outperforms SplaTAM in most metrics, demonstrating better adaptability to pseudo-depth input. The visualization shows that our method reduces trajectory jitter and maintains superior accuracy, even with pseudo-depth input. Moreover, our method achieves \textbf{over 10 times the tracking speed of SplaTAM}, and even with the addition of Local Graph Rendering (LGR), it remains nearly 5 times faster, further emphasizing its efficiency. The incorporation of LGR improves optical flow estimation, addressing issues like motion blur, changing illumination, and textureless regions, resulting in more accurate relative pose predictions.

We further evaluate the tracking performance on the TUM-RGBD dataset, where our approach also delivers competitive results compared to traditional dense RGB-D methods. However, there is still a performance gap when compared to systems that incorporate more advanced tracking techniques, such as loop closure and global optimization strategies. Overall, our method shows promising tracking performance, with a distinct advantage over existing neural-based approaches.

\begin{table}[t]
\centering
\footnotesize
\caption{\textbf{Quantitative results of view rendering  on Replica~\cite{straub2019replica}.} }
\setlength{\tabcolsep}{2pt}
\begin{tabular}{cccccccccccccc}
\toprule
\textbf{Method} & \textbf{Depth} & \textbf{Metric} & \textbf{Avg.} & \texttt{R0} & \texttt{R1} & \texttt{Of0} & \texttt{Of1} & \texttt{Of2}  \\
\midrule
\grayrowcolor  & 
& PSNR $\uparrow$        &                      24.54 &  22.39 &                      22.36 &                       27.79 &                      29.83 & 20.33 \\
\grayrowcolor Vox-Fusion & GT & SSIM $\uparrow$        &                      0.79 &                      0.68 &                      0.75 &   0.86 &                      0.88 &                      0.79 \\
\grayrowcolor & & LPIPS $\downarrow$        &                      0.25 &  0.30 &  0.27 &                                         0.24 &  0.18 &  0.24 \\
\midrule
\grayrowcolor & 
& PSNR $\uparrow$         &  24.73 &                      22.12 &  22.47 &  29.07 &  30.34 &                      19.66 \\
\grayrowcolor NICE-SLAM & GT & SSIM $\uparrow$         &   0.80 &  0.69 &  0.76 &   0.87 &  0.89 &  0.80\\
\grayrowcolor & & LPIPS $\downarrow$         &   0.25 &                      0.33 &  0.27 &   0.23 &  0.18 &  0.24  \\
\midrule
\grayrowcolor & 
& PSNR $\uparrow$ &  \textbf{35.58} &  32.40 &  \textbf{34.08} &    \textbf{38.26} &  39.16 &  \textbf{33.99}  \\
\grayrowcolor Point-SLAM & GT & SSIM $\uparrow$ &  \textbf{0.98} &  0.97 &  \textbf{0.98} &  \textbf{0.98} &    \textbf{0.99} &  0.96 \\
\grayrowcolor & & LPIPS $\downarrow$ &  0.12 &  0.12 &  0.11 &  0.10 &  0.12 &  0.16  \\
\midrule
\grayrowcolor & 
& PSNR $\uparrow$                          &  35.23 &  \textbf{32.86} &  33.89 &    \textbf{38.26} &  \textbf{39.17} &  31.97  \\
\grayrowcolor SplaTAM & GT & SSIM $\uparrow$                          &  0.98 &  \textbf{0.98} &  0.97 &  \textbf{0.98} &  0.98 &  \textbf{0.97}  \\
\grayrowcolor & & LPIPS $\downarrow$                          &  \textbf{0.09} &  \textbf{0.07} &  \textbf{0.10} &   \textbf{0.09} &  \textbf{0.09} &  \textbf{0.10} \\
\midrule
\multirow{3}{*}{SplaTAM} & \multirow{3}{*}{Pseudo} 
& PSNR $\uparrow$        &                      23.06 &  \textbf{24.82} &                      21.68 &                        23.95 &                      23.56 & \textbf{21.29} \\
& & SSIM $\uparrow$        &                      0.80 &                      \textbf{0.879} &                      0.731 &    0.764 &                      0.797 &                      \textbf{0.837} \\
& & LPIPS $\downarrow$        &                      \textbf{0.36} &  \textbf{0.259} &  0.416  &                      0.396 &  0.401 &  \textbf{0.325}  \\
\midrule
\multirow{2}{*}{Ours} & \multirow{3}{*}{Pseudo} 
& PSNR $\uparrow$         &  22.98 &                      21.35 &  \textbf{22.40} &    \textbf{25.14} &  \textbf{26.70} &  19.34  \\
\multirow{2}{*}{(w/o LGR)}& & SSIM $\uparrow$         &  0.79 &  0.777 &  \textbf{0.758} &   \textbf{0.799} &  0.865 &  0.776 \\
& & LPIPS $\downarrow$         &  0.37 &                      0.396 &  \textbf{0.403} &    \textbf{0.395} &  0.312 &  0.380  \\
\midrule
\multirow{3}{*}{Ours} & \multirow{3}{*}{Pseudo} 
& PSNR $\uparrow$ &  \textbf{22.92} &  21.54 &  22.29 &   24.87 &  26.69 &  19.23 \\
& & SSIM $\uparrow$ &  \textbf{0.79} &  0.784 &  0.753 &   0.796 &  \textbf{0.869} &  0.772 \\
& & LPIPS $\downarrow$ &  0.37 &  0.391 &  0.420 &  0.399 &  \textbf{0.306} &  0.377 \\
\bottomrule
\end{tabular}
\vspace{-3mm}
\label{tab:replica_render}
\end{table}

\begin{table}[t]
\centering
\footnotesize
\caption{\textbf{Quantitative results of view rendering on TUM-RGBD.}}
\begin{tabular}{ccccccccccccc}
\toprule
{\textbf{Method}} & {\textbf{Metric}}& {\textbf{Average}} & \texttt{fr1/desk} &  \texttt{fr1/desk2}    \\
\midrule
\multirow{3}{*}{SplaTAM}
& PSNR $\uparrow$        & \textbf{16.97} & \textbf{17.76} & 16.18  \\
& MS-SSIM $\uparrow$     & 0.660 & 0.705 & 0.614  \\
& LPIPS $\downarrow$     & 0.460 & \textbf{0.402} & 0.509\\

\midrule

\multirow{2}{*}{Ours} 
& PSNR $\uparrow$        & 16.66 & 17.06 & \textbf{16.26} \\
\multirow{2}{*}{(w/o LGR)} & MS-SSIM $\uparrow$     & \textbf{0.728} & \textbf{0.711} & \textbf{0.725}  \\
& LPIPS $\downarrow$     & \textbf{0.449} & 0.425 & \textbf{0.473} \\

\midrule

\multirow{3}{*}{Ours} 
& PSNR $\uparrow$        & 16.54 & 17.72 & 15.36 \\
& MS-SSIM $\uparrow$     &  0.684 & 0.707 & 0.660  \\
& LPIPS $\downarrow$     & 0.472 & 0.436 & 0.508 \\

\bottomrule
\end{tabular}
\vspace{-3mm}
\label{tab:TUM_render}
\end{table}

\begin{figure}[t]
    \centering
    \includegraphics[width=0.85\linewidth]{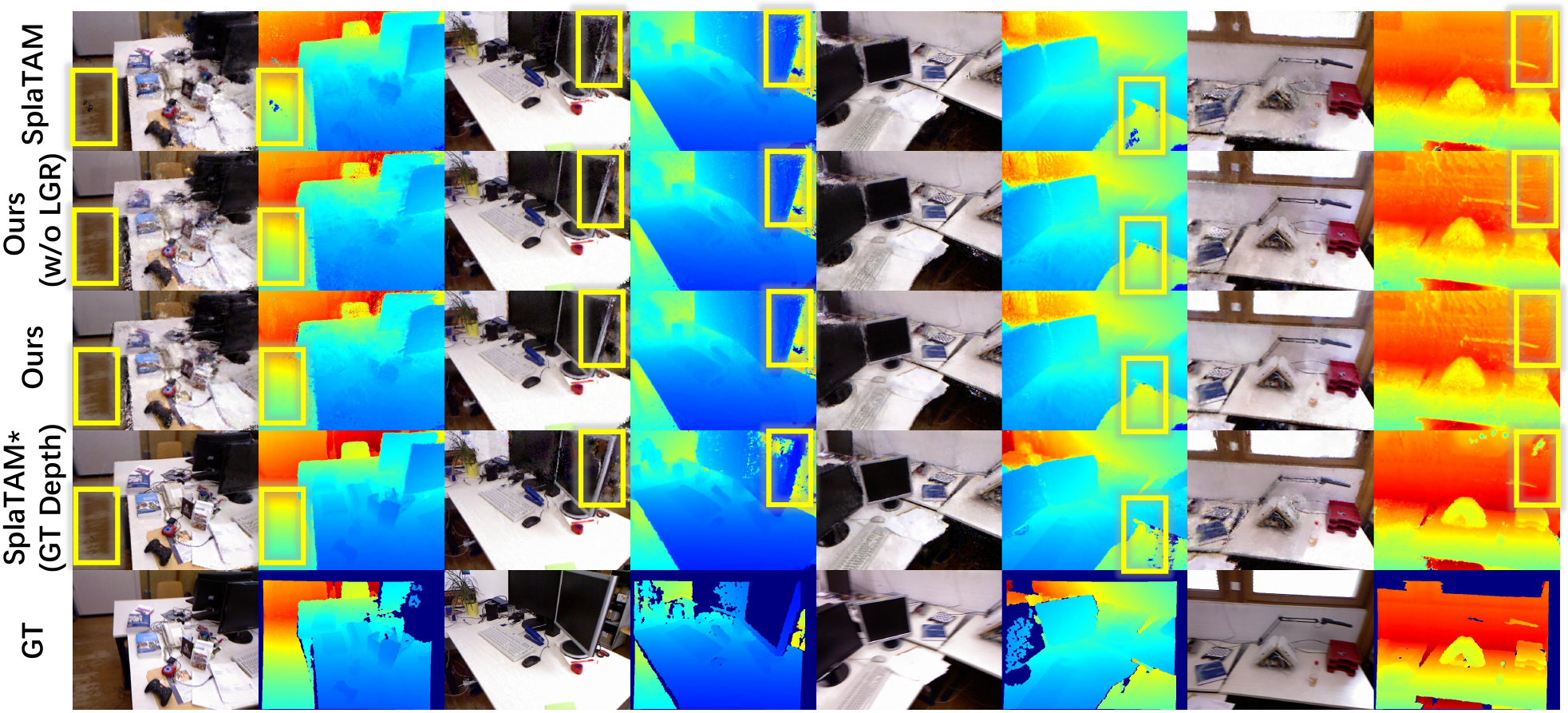} \\
    \caption{\textbf{Qualitative results on TUM-RGBD.} Our feed-forward network accelerates the model's tracking performance without compromising the quality of mapping. In fact, for some samples, we achieved better Gaussian models compared to SplaTAM, which uses GT depth as input (the 4th row).}
    \label{fig:TUM_render}
    \vspace{-4mm}
\end{figure}



\subsection{Rendering Results}
Table~\ref{tab:replica_render} and Table~\ref{tab:TUM_render} show the quantitative performance of our method compared to several baseline approaches across different datasets. 
Our method offers a significant advantage by using pseudo-depth instead of ground-truth (GT) depth, which is more realistic for real-world SLAM applications where accurate depth information is often unavailable. While many baseline methods rely on GT depth for 3D sampling and alignment, our approach achieves competitive performance even when using pseudo-depth, which is typically noisier and less reliable.
In evaluations, such as those on the Replica dataset, our method produces results close to those using GT depth, demonstrating its robustness to imperfect depth input. This highlights its potential for practical use in dynamic environments, where real-time systems must rely on estimated depth. Unlike methods that are limited by the assumption of available GT depth, our approach can generate high-quality renderings with pseudo-depth, making it more adaptable and suitable for real-world applications.


\begin{figure}[t]
    \centering
    \includegraphics[width=0.98\linewidth]{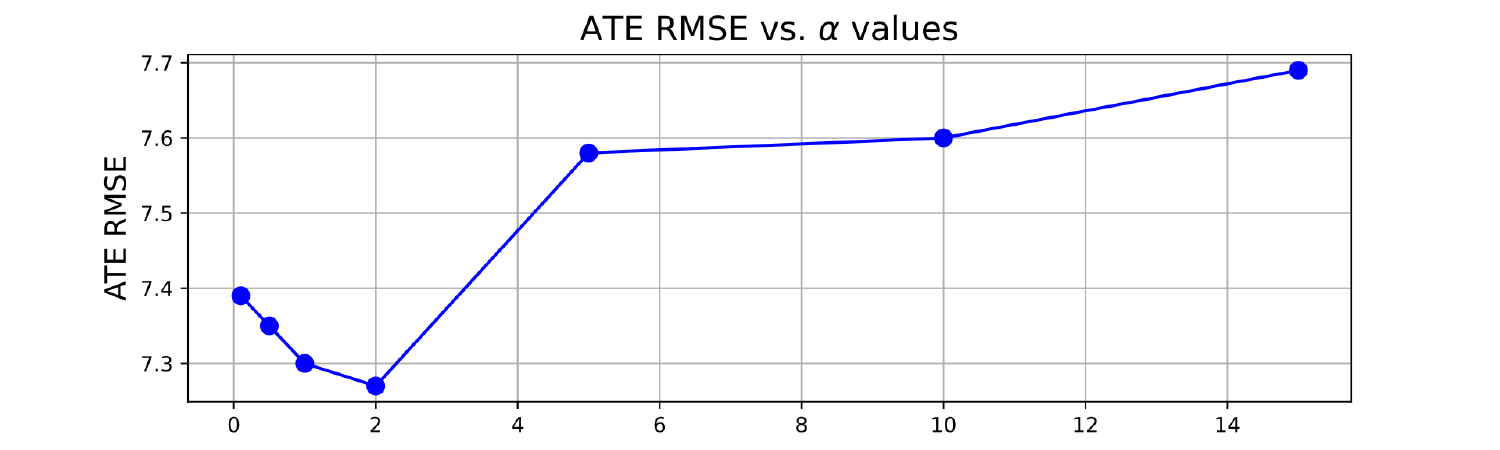} \\
    \caption{\textbf{Ablation study on the effect of the inertia coefficient $\alpha$ on performance using the TUM dataset.}}
    \label{fig:alpha}
    \vspace{-6mm}
\end{figure}

\begin{table}[!t]
\centering
\scriptsize
\setlength{\tabcolsep}{3pt}
\caption{\textbf{Runtime on Replica/R0 using an NVIDIA RTX A6000.}}
\begin{tabular}{lcccccccc}
\toprule
\multirow{2}{*}{\textbf{Method}} & \textbf{Track} & \textbf{Track} & \textbf{Map} & \textbf{Map}  & \textbf{Track} & \textbf{Map} & \textbf{ATE}\\
 & \textbf{/Iter.} & \textbf{Frame} & \textbf{/Iter.} & \textbf{/Frame} & \textbf{Total} & \textbf{Total} & \textbf{RMSE} \\
\midrule
SplaTAM~\cite{keetha2024splatam}                         &  65 ms &	2.62 s	& 78 ms &	4.69 s & 5232 s & 9384 s&	12.48  \\
Ours (w/o LGR) & None & 0.19 s &	74 ms &	4.49 s & 389 s & 8978 s&	16.74 \\
Ours & None&	0.53 s&	74 ms&	4.45 s&	1056 s & 8896 s &10.96\\
\bottomrule
\end{tabular}
\vspace{-5mm}
\label{tab:runtime}
\end{table}

\begin{table}[t]
    \footnotesize
    \centering
    \setlength{\tabcolsep}{0.03\linewidth}
    \caption{\textbf{Ablation on the graph nodes (rendering views) and sampling diversities.}}
    \vspace{-2mm}
    \begin{tabular}[b]{cccccc}
    \toprule
    \multirow{2}{*}{\textbf{$N$}}     & \textbf{$\theta$}  & \textbf{ATE} & \multirow{2}{*}{\textbf{PSNR ↑}} & \multirow{2}{*}{\textbf{SSIM ↑}} & \multirow{2}{*}{\textbf{LPIPS ↓}} \\
    & \textbf{(degree)} &  \textbf{RMSE ↓} \\
    \midrule
    \multirow{4}[1]{*}{2} & 15    & \textbf{7.39} & \textbf{15.72} & \textbf{0.672} & \textbf{0.507} \\
          & 30    & 7.45  & 15.42 & 0.663 & 0.511 \\
          & 45    & 7.51  & 15.53 & 0.67  & 0.512 \\
          & 60    & 7.58  & 15.61 & 0.67  & 0.508 \\
          \midrule
    \multirow{4}[0]{*}{4} & 15    & 7.42  & \textbf{15.92} & \textbf{0.699} & \textbf{0.485} \\
          & 30    & \textbf{7.36} & 15.76 & 0.677 & 0.504 \\
          & 45    & 7.46  & 15.48 & 0.669 & 0.506 \\
          & 60    & 7.52  & 15.57 & 0.663 & 0.514 \\
           \midrule
    \multirow{4}[0]{*}{6} & 15    & \textbf{7.31} & \textbf{15.66} & \textbf{0.660} & \textbf{0.511} \\
          & 30    & 7.50   & 15.48 & 0.653 & 0.516 \\
          & 45    & 7.51  & 15.28 & 0.658 & 0.514 \\
          & 60    & 7.50   & 15.52 & \textbf{0.660} & 0.520 \\
          \midrule
    \multirow{4}[1]{*}{8} & 15    & \textbf{7.27} & \textbf{15.91} & \textbf{0.684} & 0.502 \\
          & 30    & 7.47  & 15.68 & 0.678 & \textbf{0.500} \\
          & 45    & 7.50   & 15.67 & 0.671 & 0.509 \\
          & 60    & 7.43  & 15.44 & 0.663 & 0.517 \\
    \bottomrule
    \end{tabular}
    \label{tab:ablation}
    \vspace{-3mm}
\end{table}

\subsection{Runtime Comparison}
In Table~\ref{tab:runtime}, we compare the runtime of our method with SplaTAM~\cite{keetha2024splatam} on an NVIDIA RTX A6000 GPU. Our method offers a significant speed advantage, reducing tracking time to \(<1\)s per frame, with total times of 389s (without LGR) and 1056s (with LGR). In contrast, SplaTAM takes 65ms per iteration and 5232s for a typical sequence due to iterative pose optimization. Despite the faster runtime, our method achieves competitive accuracy with an ATE RMSE of 10.96 cm, compared to SplaTAM's 12.48 cm.
This demonstrates the effectiveness of our feed-forward pose prediction, which significantly speeds up tracking while maintaining high performance for real-time, large-scale 3D reconstruction.

\begin{figure}[t]
    \centering
    \newcolumntype{P}[1]{>{\centering\arraybackslash}m{#1}}
    \setlength{\tabcolsep}{0.001\textwidth}
    \renewcommand{\arraystretch}{0.8}
    \footnotesize
    \def\myheight{.09}
    \def\mywidth{.18}
    \begin{tabular}{c}

    \includegraphics[width=0.8\linewidth]{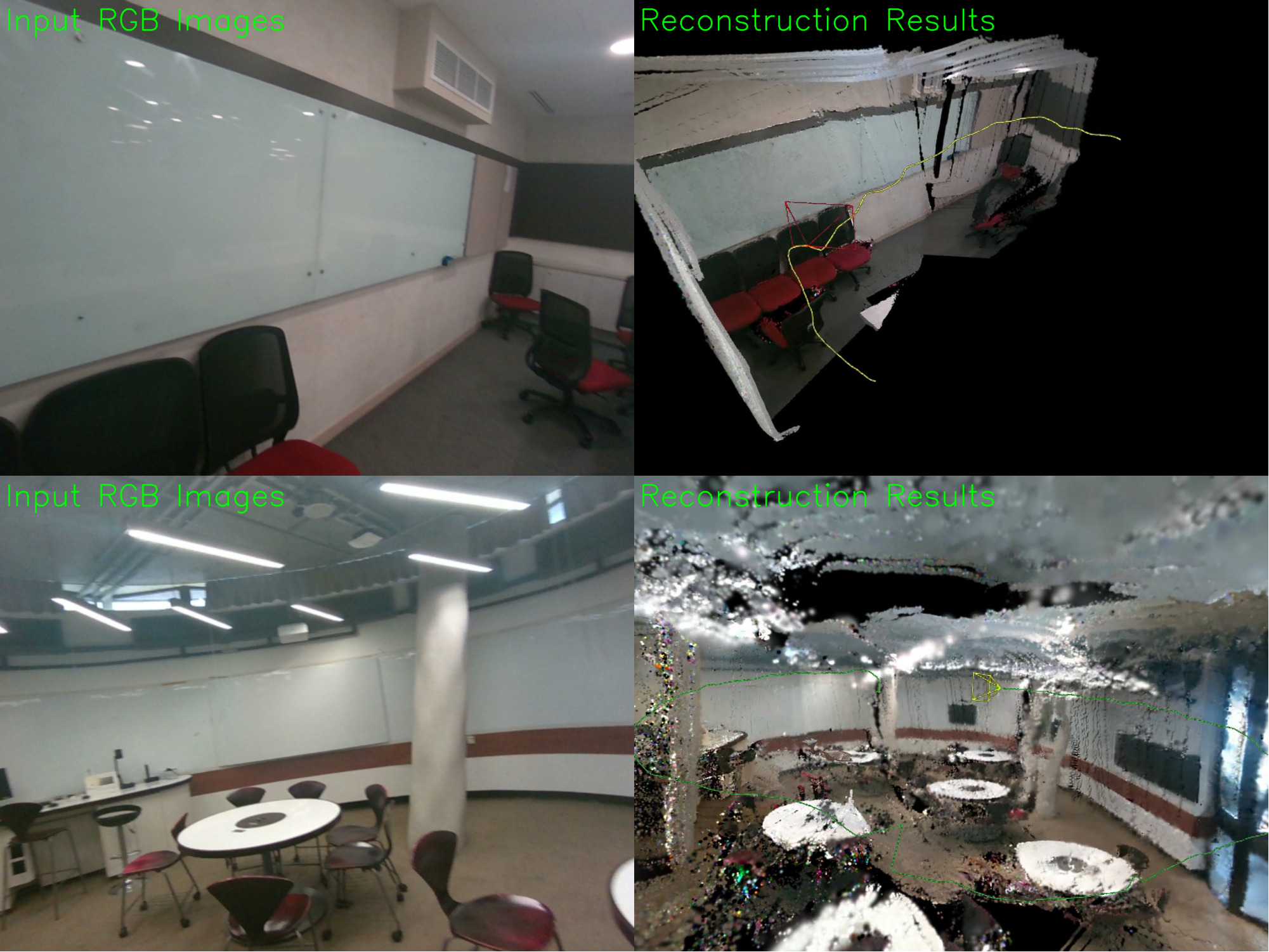} \\

    \end{tabular}
    \caption{\textbf{Screenshot of the real-world deployment demo.}}
    \label{fig:wild_demo}
    \vspace{-6mm}
\end{figure}

\subsection{Video Demonstrations}
We deployed our method on a real-world platform, where we processed two RGB video streams online: one from a conference room and the other from a classroom. Using the same experimental setup as in the main experiment, including UniDepth to generate pseudo-depth, we achieved an inference speed of over 1 fps. The visualization results are particularly impressive, as demonstrated in Fig.~\ref{fig:wild_demo}.

\subsection{Ablation Studies}
We also conducted an ablation study to evaluate the impact of the camera's inertia, the number of graph nodes, and the angular diversity of sampled views on various performance metrics, including ATE, PSNR, SSIM, and LPIPS. 
\subsubsection{\textbf{Effect of Camera's Inertia}}
We begin by examining the effect of different values for the inertia coefficient \(\alpha\). The trend observed in Fig.~\ref{fig:alpha} reveals that the value of \(\alpha\) has a relatively small effect on the ATE RMSE, with values remaining stable across different \(\alpha\) values. Specifically, for lower values of \(\alpha\) (e.g., 0.1, 0.5, 1), the ATE RMSE decreases slightly from 7.39 to 7.27. As \(\alpha\) increases (e.g., 5, 10, 15), the ATE RMSE starts to rise, peaking at 7.69 for \(\alpha = 15\). This suggests that the model performs best when \(\alpha\) is within a moderate range (e.g., 1 or 2), striking a balance between inertia and trajectory stability. Larger \(\alpha\) values appear to degrade performance, due to the increased influence of inertia, which can lead to drift by overly relying on previous estimates.

\subsubsection{\textbf{Effect of Graph Nodes}}
As shown in Table~\ref{tab:ablation}, increasing the number of graph nodes (\(N\)) from 2 to 8 improves performance. The ATE RMSE decreases, indicating better trajectory accuracy, while PSNR and SSIM improve, reflecting higher image quality. LPIPS decreases, showing better perceptual similarity to the ground truth. More nodes enhance pose prediction by incorporating additional viewpoints, boosting optical flow robustness.

\subsubsection{\textbf{Effect of Sampled Views}}
Increasing angular diversity (\(\theta\)) improves PSNR and SSIM, indicating better image quality. However, the ATE RMSE remains stable, suggesting that angular diversity mainly enhances image quality rather than trajectory accuracy. LPIPS improves with larger \(\theta\), especially at lower \(N\), reducing perceptual differences.
The best ATE RMSE performance is with \(N = 6\) and \(\theta = 15^\circ\), balancing accuracy and image quality. Higher \(N\) values like 8 show similar or slightly worse ATE. For PSNR and SSIM, \(N = 4\) and \(N = 6\) with \(\theta = 15^\circ\) perform best, while LPIPS is minimized at \(N = 4\) and \(\theta = 30^\circ\), suggesting moderate angles improve perceptual quality without sacrificing performance.


\section{Conclusion}
We propose an efficient online 3D reconstruction method that combines dense visual SLAM with 3D Gaussian mapping for scene representation. Unlike traditional methods that depend on depth sensors or slow optimization, our approach enables real-time 3D reconstruction and novel-view synthesis using only an RGB stream. We introduce a feed-forward pose prediction module, accelerating camera pose tracking through network inference instead of optimization. 
Our approach offers a promising solution for depth sensor-free, online 3D reconstruction, paving the way for future advancements in efficient SLAM systems.

\noindent \textbf{Acknowledgements}. This work was supported in part by the China Postdoctoral Science Foundation under Grant 2025M771741, in part by the Postdoctoral Fellowship Program of the China Postdoctoral Science Foundation (CPSF) under Grant GZC20251206, and in part by the National Natural Science Foundation of China under Grant U22B2050 and Grant 62376032.

\bibliographystyle{ieeetr} 
\bibliography{ref} 
\addtolength{\textheight}{-12cm}   
\end{document}